\documentclass[12pt]{article}

\usepackage[utf8]{inputenc}
\usepackage[margin=1in,footskip=0.25in]{geometry}

\usepackage{amsfonts}
\usepackage{amsmath}
\usepackage{bm}
\usepackage{indentfirst}
\usepackage{graphicx}
\usepackage{float}
\usepackage{setspace}
\usepackage{textcomp}
\usepackage{makecell}
\usepackage{multirow}
\usepackage{amssymb}
\usepackage{color}
\usepackage{subcaption}
\usepackage{authblk}
\usepackage{algorithmic}
\usepackage{algorithm2e}

\usepackage[pdftex, colorlinks, urlcolor=blue, linkcolor=blue, anchorcolor=blue, citecolor=blue]{hyperref}

\usepackage[round]{natbib}

\title{\textbf{Adapting Projection-Based Reduced-Order Models using Projected Gaussian Process}}

\author[]{Xiao Liu} 
\author[]{Jingyi Feng}
 \author[]{Xinchao Liu}
\affil[]{H. Milton Stewart School of Industrial and Systems, \\Georgia Institute of Technology}

\date{}

\begin{document}

\maketitle

\vspace{-30pt}
\begin{abstract}
Projection-based model reduction is among the most widely adopted methods for constructing parametric Reduced-Order Models (ROM). Utilizing the snapshot data from solving full-order governing equations, the Proper Orthogonal Decomposition (POD) computes the optimal basis modes that represent the data, and a ROM can be constructed in the low-dimensional vector subspace spanned by the POD basis. For parametric governing equations, a potential challenge arises when there is a need to update the POD basis to adapt ROM that accurately capture the variation of a system's behavior over its parameter space (in design, control, uncertainty quantification, digital twins applications, etc.). In this paper, we propose a Projected Gaussian Process (pGP) and formulate the problem of adapting the POD basis as a supervised statistical learning problem, for which the goal is to learn a mapping from the parameter space to the Grassmann manifold that contains the optimal subspaces. A mapping is firstly established between the Euclidean space and the horizontal space of an orthogonal matrix that spans a reference subspace in the Grassmann manifold. A second mapping from the horizontal space to the Grassmann manifold is established through the Exponential/Logarithm maps between the manifold and its tangent space. Finally, given a new parameter, the conditional distribution of a vector can be found in the Euclidean space using the Gaussian Process (GP) regression, and such a distribution is then projected to the Grassmann manifold that enables us to predict the optimal subspace for the new parameter. As a statistical learning approach, the proposed pGP allows us to optimally estimate (or tune) the model parameters from data and quantify the statistical uncertainty associated with the prediction. The advantages of the proposed pGP are demonstrated by numerical experiments.  
\end{abstract}

\newpage

\singlespacing

\section{Introduction} \label{sec:intro}
\subsection{Background and Problem Statement} \label{sec:overview}
Consider the diffusion of chemical species, aeroelasticity of aircraft wings, flow of vehicles and predator-prey interactions, dynamics of such complex systems is typically described by a set of Partial Differential Equations (PDEs). Very often, the dimensions of these problems are  high after the spatial discretization of the mathematical models (e.g., computational fluid dynamics models or finite element analysis with very large numbers of degrees of freedom). 
The prohibitive computational cost prevents us from repeatedly running numerical solvers under  different parameter settings; often needed for design, diagnosis, prediction, uncertainty quantification and digital twins applications \citep{Peherstorfer2015,Peherstorfer2015b,Sargsyan2015}. 
Hence, when there exist low-dimensional patterns embedded in high-dimensional systems \citep{kutz2016dynamic}, Reduced-Order Models (ROM) become indispensable in capturing the dominate behaviors of complex systems. Compared with data-driven surrogate models or meta models \citep{Russell2006}, ROM often preserve  important system physics, such as stability \citep{Prajna2003,Batool2021}, passivity \citep{Sorensen2003,Breiten2022}, polynomial nonlinearities \citep{Qian2020}, etc. 

Among various model reduction approaches, projection-based model reduction is perhaps the most widely adopted one for constructing ROM \citep{kutz2016dynamic, Qian2020}. 
The central idea behind projection-based ROM is to project a high-dimensional system to an optimal low-dimensional vector space spanned by orthogonal bases. Although searching for such bases may seem to be a nontrivial task considering the computational speed, accuracy and complex boundary conditions, the Proper Orthogonal Decomposition (POD)---applied to curves in Hilbert spaces $\mathcal{H}$ of infinite dimension---provides a computationally efficient approach that optimally approximates data generated from full-order systems in an $\ell_2$ sense \citep{berkooz1993proper, rowley2005model, Benner2015}.
This can be efficiently done by performing the thin Singular Value Decomposition (SVD) of the snapshot data matrix, and one may refer to  \cite{kutz2016dynamic} for a more detailed description of the procedure.

\vspace{6pt}
To elaborate, consider a governing equation with parameters $\bm{\theta}=(\theta_1, \theta_2, ..., \theta_d) \in  \mathbb{P} \subset \mathbb{R}^d$, where $\mathbb{P}$ is the parameter space. Given $\bm{\theta}$, the solution of the system, $x(t,\bm{s};\bm{\theta})$, is a space-time field, 
\begin{equation}
(t, \bm{s}) \in [0,T] \times \mathbb{S} \mapsto x(t,\bm{s};\bm{\theta})
\end{equation}
where $\mathbb{S}$ is the spatial domain and $T>0$. 
Let $\bm{x}(t;\bm{\theta})=(x(t,\bm{s}_1;\bm{\theta}),x(t,\bm{s}_2;\bm{\theta}),\cdots, x(t,\bm{s}_n;\bm{\theta}))$ be an $n$-dimensional vector representing the discretized solutions at a set of locations $\{\bm{s}_1, \bm{s}_2, \cdots, \bm{s}_n\}$,  one obtains the \textit{snapshot} data matrix, $\bm{D}(\bm{\theta}) = [\bm{x}(t_1;\bm{\theta});\bm{x}(t_2;\bm{\theta});\cdots;\bm{x}(t_{n_T};\bm{\theta})]$ with $n_T$ being the number of temporal points.   
The POD method finds the optimal $r$-dimensional ($r \ll n$) vector subspace $\mathbb{W} \subset \mathbb{R}^n$  and an orthogonal projection, $\mathcal{W}:\mathbb{R}^n \rightarrow \mathbb{W}$, such that the following distance is minimized
\begin{equation}
\label{eq:POD_minimization_1}
\text{min}_\mathcal{W}\sum_{i=1}^{n_T}||\bm{x}(t_i;\bm{\theta})-\mathcal{W}(\bm{x}(t_i;\bm{\theta}))||^2
\end{equation}
where $||\cdot||=\sqrt{\left \langle \cdot,\cdot \right \rangle}$ and $\left \langle \cdot,\cdot \right \rangle$ is the inner product of the Hilbert space. 

In particular, let $\bm{\Phi}= (\bm{\phi}_1,\bm{\phi}_2,\cdots,\bm{\phi}_{r}) \in \mathbb{R}^{n\times r}$ be a matrix basis, with orthogonal columns and $\bm{\Phi}^T\bm{\Phi}=\bm{I}_r$, that spans a  subspace $\mathbb{R}^r$ such that $\bm{x}(t;\bm{\theta}) \approx \bm{\Phi}\bm{x}_r(t;\bm{\theta})$. The matrix basis $\bm{\Phi}$ can be found by $\mathrm{min}_{\bm{\Phi}}||  \bm{D}(\bm{\theta})- \bm{\Phi}\bm{\Phi}^T \bm{D}(\bm{\theta})||^2_F$, 
where $||\cdot||_F$ is the Frobenius norm of matrices in the vector space. 
The Eckart-Young theorem \citep{Eckart1936} shows that the solution of the minimization problem  above is obtained by the thin SVD of the snapshot matrix $\bm{D}(\bm{\theta})$, i.e., $\bm{\Phi}=[\bm{u}_1, \bm{u}_2, \cdots, \bm{u}_r]$ retains the leading $r$ columns of $\bm{U}$ from $\bm{D}(\bm{\theta})=\bm{U}\bm{\Sigma}\bm{V}^T$. 
For the approach outlined above, it is seen that the optimal POD basis---obtained from the snapshot data generated at the parameter setting $\bm{\theta}$---is only optimal to represent the data generated from that parameter setting. 

\vspace{6pt}
Hence, for parametric dynamical systems, 
it is critical for the POD modes to be sufficiently representative so that data generated from various parameter settings are well represented. This is particularly important for design, diagnosis, control, uncertainty quantification and real-time operations (e.g., digital twins) where parameter changes are common \citep{Amsallem2009, Peherstorfer2015, Peherstorfer2015b, Sargsyan2015}. 
One approach, known as the greedy basis generation, finds the reduced basis by an iterative procedure   \citep{Hesthaven2016}. At each iteration, one new basis function is added using one solution computed at a new parameter setting. Throughout multiple iterations, the overall precision is gradually enhanced and the final basis is rich enough to represent data from the parameter space. 

%

\begin{figure}[h!]
\includegraphics[width=1.0\textwidth]{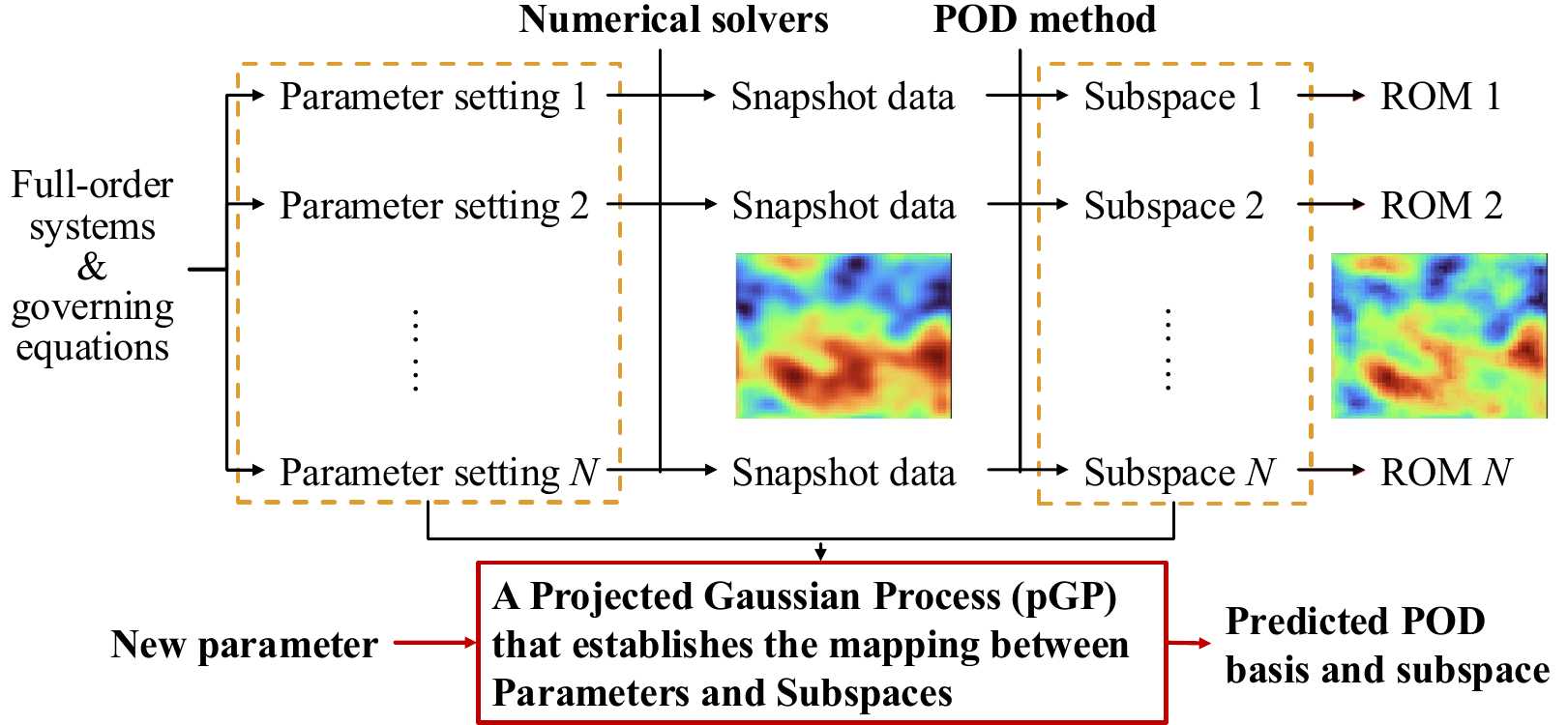}
\caption{A supervised statistical learning problem based on the proposed pGP that enables the prediction of POD basis modes as parameters change.}
\label{fig:overview}
\end{figure}

Alternatively, the optimal basis can be adapted as parameters change so that the POD basis can capture the variation of a system's behavior over its parameter space  \citep{Bui2008, Vetrano2011}. 
In this paper, we focus on this approach and formulate the problem of adapting the POD basis as a supervised statistical learning problem. The goal is to learn a mapping between the parameter $\bm{\theta}$ and POD basis $\bm{\Phi}$ that spans a low-dimensional vector subspace. Such a subspace regression problem is stated as follows: 

\vspace{6pt}
\textit{Given $\{\bm{\theta}_i, \bm{\Phi}_i\}_{i=1}^k$ for $\bm{\theta}_i \in \mathbb{P}$, the problem is to learn a mapping $\mathcal{T}: \mathbb{P}  \rightarrow  \mathcal{ST}(r,n)$ from the parameter space $\mathbb{P} \subset \mathbb{R}^d$ to a compact Stiefel manifold of matrices with orthogonal columns, which enables the prediction of the POD basis matrix $\bm{\Phi}^*$ given a new parameter $\bm{\theta}^*$ without numerically solving the computationally expensive full-order governing equation at $\bm{\theta}^*$ (see Figure \ref{fig:overview}).} 

\vspace{6pt}
We address this problem by proposing a supervised statistical learning approach, known as the Projected Gaussian Process (pGP) regression. As shown in Figure \ref{fig:overview}, when (training) data are available by numerically solving the full-order governing equation repeatedly at a (usually small) number of parameter settings $\{\bm{\theta}_{i}\}_{i=1}^k$, it is possible to obtain the optimal basis $\{\bm{\Phi}_{i}\}_{i=1}^k$ at each parameter setting, separately, using the existing approach. Once the proposed pGP regression learns the mapping between parameters and POD bases, a new basis can be predicted given the new parameters, thus adapting the ROM for parameters change (see Sec.\ref{sec:GP}). 

\subsection{Literature Review} 

Various approaches have been investigated for tackling the model reduction problem of parametric dynamical systems. Some examples include the interpolatory model reduction and rational interpolation  \citep{Baur2011, Gosea2021}, multi-moment matching  \citep{Freund2003, Benner2014book}, balanced truncation  \citep{Wittmuess2016}, matrix interpolation  \citep{Panzer2010, Geuss2013}, etc. For a comprehensive and complete review of projection-based model reduction for parametric dynamical systems, we refer the readers to \citep{Benner2015, Benner2017Review}. 

\begin{figure}[h!]
\includegraphics[width=1.0\textwidth]{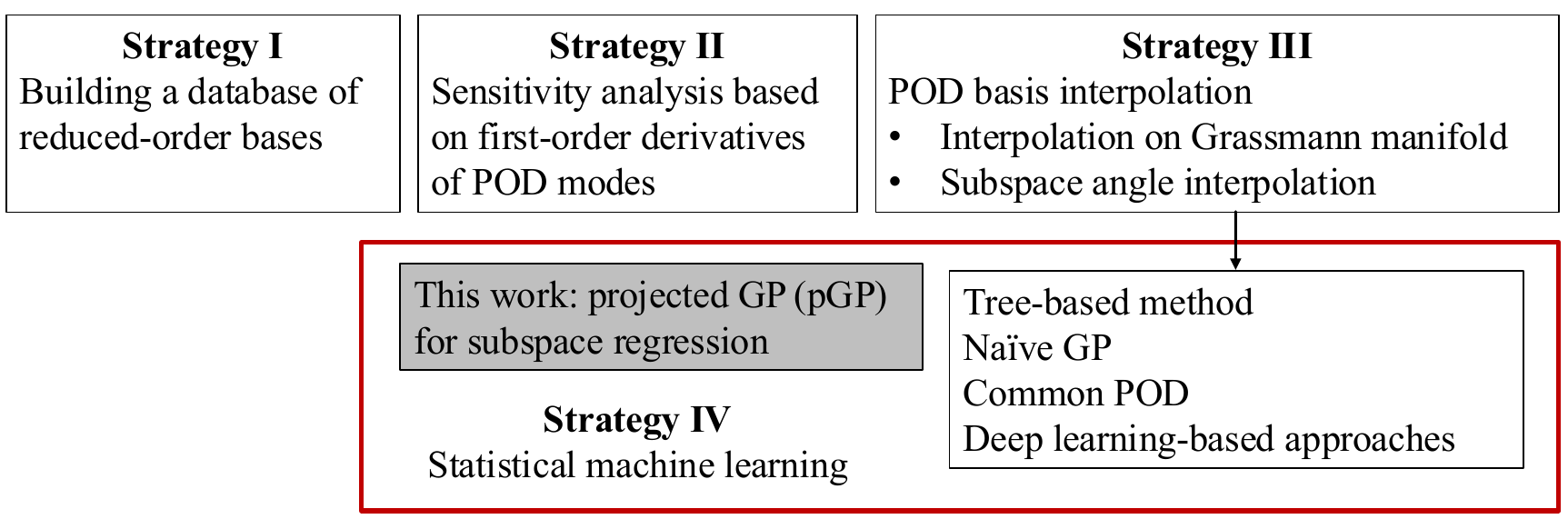}
\caption{Existing strategies to adapt ROM for parameter changes}
\label{fig:literature}
\end{figure}

\vspace{6pt}
In this section, we primarily focus on major strategies  for adapting the POD basis for parameters change. Figure \ref{fig:literature} presents one possible way to group existing approaches into four categories. This figure is only used to facilitate the explanations of different approaches as well as how the proposed approach is built upon the prior work. The boundaries that separate different approaches may change if we examine these methods from different perspectives. 

$\bullet$ Strategy I involves pre-computing a database, i.e., a library, of reduced-order bases \citep{Amsallem2009, Kapteyn2020}. For applications where this approach is feasible, ROM can be updated in real time over the set of parameters included in the database. 

$\bullet$  Strategy II computes the first-order total derivatives of the POD modes with respective to a parameter \citep{Borggaard2008}, and a new POD basis, given a small change of the parameters from a base point, can be found through the first-order expansion in the parametric space. 

$\bullet$ Strategy III involves different approaches that aim to interpolate or predict new POD bases from the observed parameter-POD basis pairs. 
In particular, the POD basis interpolation on Grassmann manifolds have been extensively investigated \citep{Amsallem2008, Son2013, Benner2015, Mosquera2019}. This approach is based on an important result that a subspace spanned by a POD basis is a point on the Grassmann manifold that consists of $r$-dimensional subspaces in $\mathbb{R}^n$, where $n$ and $r$ are respectively the original and reduced-order dimensions. Hence, a tangent space, attached to a given base point on the Grassmann manifold, can be constructed, and the subspaces from the Grassmann manifold can be projected to the tangent space through a \textit{logarithmic} map. After this critical step, interpolation can be done on the tangent space (such as the Lagrangian interpolation \citep{Amsallem2008}, inverse distance weighting \citep{Mosquera2019}, or Gaussian Process \citep{Giovanis2020}), and the interpolated vector on the tangent space is eventually mapped back to the Grassmann manifold through an \textit{exponential} map. Hence, this method depends on the choice of the base point upon which the tangent space is attached to, and naturally requires the injective radius of the exponential map to be sufficiently large so that the approach is stable \citep{Friderikos2022}. In contrast, the subspace angle interpolation approach directly computes the geometric distance between two subspaces by finding the principal angels between two POD bases \citep{Ye2016}. The principal angels can be viewed as a series of angles required to rotate one subspace to form another, and such a distance metric is used to interpolate subspaces. Other examples based on the idea of interpolation include the matrices interpolation  of local reduced models for linear time-invariant systems  \citep{Panzer2010}, and  \cite{Son2013} investigated the interpolation on Grassmann manifolds for affinely dependent system matrices on model parameters. 

\vspace{-12pt}
$\bullet$ Strategy IV involves using statistical machine learning to learn the mapping between parameters and POD bases (or the subspaces spanned by POD bases). Leveraging the key ideas from both the POD basis interpolation on Grassmann manifold \citep{Amsallem2008} and subspace angle interpolation \citep{Ye2016}, a regression tree-based approach on Grassmann manifold was proposed in  \cite{Liu2023}. Following this approach, a tree is grown by repeatedly splitting the tree node (i.e., a binary partition the parameter space) to maximize the Riemannian distance between two subspaces (based on the principals angles) spanned by the predicted POD bases on the left and right daughter nodes. As a result, this tree-based approach effectively divides the parameter space into sub-regions, and the POD basis for a sub-region is obtained using only the data generated from that sub-region. This approach successfully combines the POD basis prediction problem into the classical framework of regression trees, and appears to be more stable and conservative than the existing POD basis interpolation on Grassmann manifolds (as the tree-based method only updates the POD basis when the parameter jumps from one sub-region to another in the parameter space). We also note that, this approach is useful for finding the POD basis when the dimension of the parameter space is high and the system behaviors have different sensitivity levels against different parameters. For those less important parameters, the tree-based approach effectively exclude those parameters from being used to split the tree (note that, the existing interpolation method does not automatically do this). Finally, in very recent years, neural networks have also been used for adapting ROM. For example, \cite{franco2024deep} proposed the Deep Orthogonal Decomposition by constructing a deep neural network to approximate the solution manifold through a continuously adaptive local basis (i.e., use a neural network to learn the mapping from parameters and subspaces). \cite{berman2024colora} proposed to continuously adapt the low-rank weights of pre-train neural networks to predict the evolution of solution fields at new physics parameters. In general, training a neural network requires a large amount of training data (e.g., \cite{franco2024deep} used 500 training samples for model reduction of the Eikonal equation). This may potentially pose additional challenges when it is expensive to generate training data by solving full-order physics. 

Other statistical learning approaches involve the use of GP.  For example, \cite{Giovanis2020} directly used GP to capture the relationship between parameters and the matrices that determine the vectors on the tangent space of the Grassmann manifold. Hence, given a new parameter setting, GP is used to predict a new vector in the tangent space, which yields the predicted subspace on the Grassmann manifold through an exponential map. Note that, because GP performs prediction through a weighted linear combination of responses, this approach is similar to the existing Lagrangian POD basis interpolation on Grassmann manifolds but essentially with different interpolation weights.  
For the reasons noted in Sec.\ref{sec:GP}, we refer to this approach as the native GP approach following the terminology in  \cite{Mallasto2018}.  \cite{Mak2018} proposed another approach known as the Common POD (cPOD). In an engine injector design example, a common POD basis is found for some base design geometry, the POD bases for other geometries are obtained by rescaling the common POD through a linear map. This approach requires sound domain knowledge of the system under consideration to determine the base design and establish the linear map between POD bases from the common POD.  \cite{Zhang2022} proposed a GP subspace regression where the central idea is to directly construct a matrix-variate GP for the matrix bases such that the approach does not hinge on the differential geometric structures of the Gassmann manifold.

\subsection{Contributions of this Paper} 

The prior work in the literature has shown the great potential of leveraging machine learning for adapting projection-based ROM. 
The main contribution of this paper is to propose a new supervised statistical learning method that enables the learning of the mapping between parameters and POD bases. To our best knowledge, this pGP approach is \textit{not yet available in the literature}. 

\vspace{6pt}
As a statistical machine learning approach, the proposed pGP provides some critical capabilities much needed for adapting ROM against parameter changes:

\textit{i}). Unlike existing interpolation methods (e.g., Lagrangian interpolation), the proposed statistical learning approach allows us to optimally estimate (or tune) model parameters based on training data (i.e., the predictions are data-driven). In contrast, when the Lagrangian interpolation is used, the Lagrange bases are computed from pre-existing equations which are independent of the problem or data of interest. Because the pGP model is always trained and optimized given a training dataset, such a flexibility provides the opportunity to further improve the accuracy of the predicted POD basis modes for specific problems at hand (as shown in the numerical examples)

\textit{ii}). The proposed pGP not only provides the point prediction of the POD basis, but also quantifies the statistical prediction uncertainty associated with the predicted POD modes. Such a capability of uncertainty quantification facilitates decision-making under uncertainty, and provides a critical link (as a future research path) to a large body of literature on statistical experimental design and deep reinforcement learning for more efficient construction of the POD basis by sequentially exploring the solution space of full-order models \citep{Ivanova2021, Blau2022}.

\textit{iii}). The proposed pGP, as a statistical learning approach, provides the capability of parameter selection when constructing the mapping between parameters and POD bases. This capability is needed when not all parameters are equally important in terms of their predictive power for POD modes. If important parameters can be identified, the statistical pGP model may require a smaller number of input. When experimental design is used to determine the optimal parameter settings for subsequent experiment, this can also potentially lower the dimension of the parameter space (i.e., the decision space in experimental design becomes smaller). We would also like to point out that this argument shares some similarities with the concept of active subspace  \citep{Constantine2014}. The active subspace is usually a low-dimensional space of physics model input which contains the detected directions of the strongest variability in model output, while the parameter selection mentioned in this paper refers to the process of identifying the parameters which have stronger predictive power of the POD modes in a statistical model. Hence, the outcome of identifying the active subspace allows one to construct an approximation of the physics model on a low-dimensional subspace, while the outcome of parameter selection in this paper allows one to predict the POD modes using a smaller set of parameters. 

\textit{iii}). Due to the parameter estimation/selection capabilities above, the pGP approach naturally involves the use of global POD basis as its special case. As shown in this paper, it is possible to set the parameters in the proposed pGP regression model such that the predicted POD basis matrices for different parameters become \textit{i.i.d.} samples from a matrix-variate Gaussian distribution with the global POD basis matrix being the mean of this distribution (as shown in the numerical examples).

\vspace{6pt}

The paper is organized as follows. In Section 2, we provide the preliminaries on POD basis interpolation. The proposed pGP approach is presented in Section 3. Numerical examples and discussions are presented in Section 4. Section 5 concludes the paper.

\vspace{6pt}
\section{Preliminaries on POD basis interpolation} \label{sec:preliminaries}
This section provides preliminaries on the interpolation of POD bases. Suppose that the snapshot data matrices are generated at a set of parameter settings, $\bm{\Lambda}=\{\bm{\theta}_1, \bm{\theta}_2, \cdots, \bm{\theta}_k\}$, and the optimal POD bases $\bm{\Phi}_1, \bm{\Phi}_2, \cdots, \bm{\Phi}_k$ are obtained for each parameter in $\bm{\Lambda}$. 
Then, the POD basis interpolation problem involves finding the POD basis $\bm{\Phi}^*$ at new parameters $\bm{\theta}^*\notin  \bm{\Lambda}$. Some important concepts are firstly introduced. 

\vspace{3pt}
\textbf{Grassmann manifold.} In honour of Hermann Grassmann, a Grassmann manifold, $\mathcal{G}(r,n):=\{\mathbb{W}\subset \mathbb{R}^n, \mathrm{dim}(\mathbb{W})=r\}$, is a differentiable manifold that collects all $r$-dimensional linear subspaces in $\mathbb{R}^n$ \citep{Bendokat2020}. Hence, the subspace $\mathbb{W}$ is a point in $\mathcal{G}(r,n)$
and the optimization problem in (\ref{eq:POD_minimization_1}) essentially seeks the optimal point in $\mathcal{G}(r,n)$ that minimizes the distance $\sum_{i=1}^{n_T}||\bm{x}(t_i;\bm{\theta})-\mathcal{W}(\bm{x}(t_i;\bm{\theta}))||^2$. 

\vspace{3pt}
\textbf{Stiefel manifold and a submersion.} A subspace in $\mathcal{G}(r,n)$ is spanned by an orthogonal basis $\bm{\Phi}\in \mathbb{R}^{n\times r}$ ($\bm{\Phi}^T\bm{\Phi}=\bm{I}_r$). A collection of $n\times r$ matrices with orthogonal columns forms a compact Stiefel manifold denoted by $\mathcal{ST}(r,n)$. Note that, if $\bm{\Phi}$ spans a subspace $p \in \mathcal{G}(r,n)$, so does $\bm{\Phi}\bm{J}$ for any $r\times r$ matrix $\bm{J}$ such that $\bm{J}^T\bm{J}=\bm{I}_p$. Therefore, there exists a fiber bundle
\begin{equation}\label{eq:fiber}
\pi: \bm{\Phi} \in \mathcal{ST}(r,n) \mapsto \pi(\bm{\Phi}) = p \in \mathcal{G}(r,n)
\end{equation}
which indicates that any point $p \in \mathcal{G}(r,n)$ can be represented by a point of the fiber $\pi^{-1}(p)$ \citep{Friderikos2022}. 

\vspace{3pt}
\textbf{Riemannian metric.} To establish the Riemannian metric on $\mathcal{G}(r,n)$, a unique tangent space $\mathcal{T}_p\mathcal{G}(r,n)$ is attached to a point $p \in \mathcal{G}(r,n)$. The tangent space $\mathcal{T}_p\mathcal{G}(r,n)$ has the same dimension of $\mathcal{G}(r,n)$ (but is isomorphic to $\mathbb{R}^{r\times (n-r)}$, i.e., $\mathcal{T}_p\mathcal{G}(r,n) \simeq \mathbb{R}^{r\times (n-r)}$) and is equipped with a scalar product. For any two velocity vectors, $v_1, v_2 \in \mathcal{T}_p\mathcal{G}(r,n)$, we have 
\begin{equation}
    \left \langle v_1,v_2 \right \rangle := \left \langle \bm{Z}_1,\bm{Z}_2 \right \rangle = \mathrm{trace}(\bm{Z}_1^T \bm{Z}_2)
\end{equation}
where $\bm{Z}_1$ and $\bm{Z}_2$ are $n\times r$ matrices from the horizontal space of $\bm{\Phi} \in \pi^{-1}(p)$ denoted by $\mathbb{H}_{\bm{\Phi}}:=\{\bm{Z}\in \mathbb{R}^{n\times r}; \bm{Z}^T\bm{\Phi}=0\}$. In other words, given a base point $p$, the Grassmann manifold inherits the Riemannian structure of the Stiefel manifold due to (\ref{eq:fiber}) and is a Riemannian manifold. 

\vspace{3pt}
\textbf{Exponential and Logarithm maps.} For any point $p \in \mathcal{G}(r,n)$ and the attached tangent space $\mathcal{T}_p\mathcal{G}(r,n)$, the Exponential map is defined as
\begin{equation} \label{eq:ExpMap}
\text{Exp}_p:  v \in \mathcal{T}_p\mathcal{G}(r,n) \mapsto \pi(\bm{\Phi} \bm{V}_{\bm{Z}}\cos(\bm{\Sigma}_{\bm{Z}}) + \bm{U}_{\bm{Z}}\sin(\bm{\Sigma}_{\bm{Z}}))\in\mathcal{G}(r,n)
\end{equation}
where $\bm{\Phi} \in \pi^{-1}(p)$ and $\bm{Z}=\bm{U}_{\bm{Z}}\bm{\Sigma}_{\bm{Z}}\bm{V}_{\bm{Z}}$ is the SVD of $\bm{Z} \in \mathbb{H}_{\bm{\Phi}}$. Note that, for a given $\Phi$ such that $\pi(\Phi)=p$, the tangent space $\mathcal{T}_p\mathcal{G}(r,n)$ is isomorphic to the horizontal space $\mathbb{H}_{\bm{\Phi}}$. For any $v \in \mathcal{T}_p\mathcal{G}(r,n)$, there is a unique $\bm{Z} \in \mathbb{H}_{\bm{\Phi}}$ known as the horizontal lift of $v$ \citep{Friderikos2022}. 

For any point $p \in \mathcal{G}(r,n)$ and a open set $U_{\bm{p}}:=\{p' \in \mathcal{G}(r,n)\}$, the Logarithm map is defined as
\begin{equation} \label{eq:LogMap}
    \text{Log}_p: p' \in U_{\bm{p}} \mapsto \text{Log}_p(p') \in \mathcal{T}_p\mathcal{G}(r,n). 
\end{equation}

\vspace{3pt}
\textbf{Existing POD interpolation method.} Within $U_{\bm{p}}$, the Logarithm map $\text{Log}_p$ is a diffemorphism, and the one-to-one correspondence between $v \in \mathcal{T}_p\mathcal{G}$ and $p' \in U_{\bm{p}}$ can be defined through the Exponential and Logarithm maps \citep{Friderikos2022}. Hence,  \cite{Amsallem2008} described a popular Lagrangian interpolation method for POD bases. In a nutshell, given the training data $\{\bm{\theta}_i, \bm{\Phi}_i\}_{i=1}^k$,  subspaces corresponding to $\bm{\theta}_i$ are mapped to vectors $v_i$ in the tangent space through the Logarithm map. Then, the interpolation of vectors can be done in the tangent space, and the interpolated vector $v^*$ is mapped back to the Grassmann manifold through the Exponential map. 
Algorithm 2.1 provides an example of such an approach for the one-dimensional case, i.e., $\theta \in \mathbb{R}$.

\begin{algorithm}
\caption{POD basis interpolation on Grassmann manifold}
\label{alg:POD_interpolation}
\begin{algorithmic}
\STATE{Input: Integers $r$ and $n$ ($2r\leq n$), $\{\theta_i, \bm{\Phi}_i\}_{i=1}^k$, and a target parameter $\theta^*$}. 

\STATE{ 
\begin{itemize}
\item Choose a matrix $\bm{\Phi}_0 \in \{ \bm{\Phi}_1, \bm{\Phi}_2, \cdots, \bm{\Phi}_k \}$ such that $\bm{\Phi}_0^T\bm{\Phi}_i$ is non-singular for all $i$.
\item For each $i=1,2,\cdots,k$, perform the thin SVD on $\bm{\Phi}_i(\bm{\Phi}_0^T\bm{\Phi}_i)^{-1}-\bm{\Phi}_0=\bm{U}_i\bm{\Sigma}_i\bm{V}_i^T$, and obtain a matrix $\bm{Z}_i$
\begin{equation} \label{eq:Z}
\bm{Z}_i = \bm{U}_i \text{arctan}(\bm{\Sigma}_i)\bm{V}_i^T.
\end{equation}
\item Compute an interpolated matrix and a thin SVD
\begin{equation} \label{eq:Lagrangian}
\bm{Z}^*= \sum_{i=1}^{k}\prod_{i \neq j} \frac{\theta^*-\theta_j}{\theta_i-\theta_j}\bm{Z}_i=\bm{U}^*\bm{\Sigma}^*\bm{V}^*.
\end{equation}
Note that, when the dimension of the parameter space $d$ is larger than one, the Lagrangian interpolation in (\ref{eq:Lagrangian}) can be replaced by other multidimensional interpolation methods. 
\item Return an instability message if the largest singular value of $\bm{Z}^*$ is greater than $\pi/2$. Otherwise, compute a $n\times r$ matrix
\begin{equation}
\bm{\Phi}^*= \bm{\Phi}_0\bm{V}^*\text{cos}\bm{\Sigma}^* + \bm{U}^*\text{sin}\bm{\Sigma}^* 
\end{equation}
where sin and cos only act on the diagonal of $\bm{\Sigma}^*$. 
\end{itemize}
}
\RETURN The interpolated POD basis $\bm{\Phi}^*$ for parameter $\theta^*$. 
\end{algorithmic}
\end{algorithm}

\section{Projected Gaussian Process for Adapting ROM} \label{sec:GP}
In this section, we present a supervised statistical learning method, based on the proposed pGP, for adapting POD bases against parameters change. 

\subsection{A projection from $\mathbb{R}^{nr-r}$ to $\mathcal{G}(r,n)$}
We first show that there exists an injective mapping from the Euclidean space $\mathbb{R}^{nr-r}$ to the Grassmann manifold $\mathcal{G}(r,n)$. This result will later enable us to define a GP on $\mathbb{R}^{nr-r}$ and map it to $\mathcal{G}(r,n)$.

From the preliminaries in Sec.\ref{sec:preliminaries}, there is an isomorphism between the horizontal space $\mathbb{H}_{\bm{\Phi}}$ and the tangent space $\mathcal{T}_p\mathcal{G}(r,n)$, i.e., $\pi_{\mathbb{H}}: \mathbb{H}_{\bm{\Phi}} \rightarrow  \mathcal{T}_p\mathcal{G}(r,n)$. Hence, 
it is possible to find a unique $\bm{Z} \in \mathbb{H}_{\bm{\Phi}}$ such that $\pi_{\mathbb{H}}(\bm{Z}) = v \in  \mathcal{T}_p\mathcal{G}(r,n)$, 
and the matrix $\bm{Z}$ is called the \textit{horizontal lift} of the vector $v \in  \mathcal{T}_p\mathcal{G}(r,n)$. 
The existence of such an isomorphism may suggest that a matrix-variate GP can be employed to model the relationship between $\bm{\theta}$ and the matrix $\bm{Z}$. However, \textit{it is important to point out that this seemingly natural approach needs to be carefully carried out because the matrix $\bm{Z}$ should be defined on the horizontal space $\mathbb{H}_{\bm{\Phi}}$ given a base point $p$, rather than the Euclidean space $\mathbb{R}^{n\times r}$}. Precisely speaking, if $\bm{Z}$ is modeled as a GP with respect to the parameter $\bm{\theta}$ on $\mathbb{R}^{n\times r}$, then, the point prediction, $\hat{\bm{Z}}^*$, is given by a linear combination of $\bm{Z}_1,\bm{Z}_2,\cdots,\bm{Z}_k$ based on the well-known results of GP regression \citep{Rasmussen2005}. Since $\bm{Z}_1,\bm{Z}_2,\cdots,\bm{Z}_k$ are in the horizontal space $\mathbb{H}_{\bm{\Phi}}$ , the point prediction $\hat{\bm{Z}}^*$ remains in the same space. However, the uncertainty quantification or predictive internal of $\hat{\bm{Z}}^*$ (or, the predicted $\hat{\bm{\Phi}}^*$) is  problematic because a sample $\bm{Z}$ from a Gaussian distribution on $\mathbb{R}^{n\times r}$ does not necessarily belong to $\mathbb{H}_{\bm{\Phi}}$.


\vspace{6pt}
To remedy this issue, we leverage the following result:

Proposition 3.1. \label{prop:1}
For a given basepoint $p \in \mathcal{G}(r,n)$ and its corresponding basis matrix $\bm{\Phi}=[\bm{\phi}_{1},\cdots,\bm{\phi}_{r}]$, i.e, $\pi(\bm{\Phi})=p$, it is possible to obtain  a $n\times r$ matrix $\bm{Z}$ from a $(nr-r)\times nr$ matrix $\bm{F}$  and a $(nr-r)\times 1$ vector $\bm{y}$ as follows
\begin{equation} \label{eq:y_to_Z}
\bm{Z} = \mathrm{Mat}_{n, r}(\bm{F}^{T}\bm{y})
\end{equation} 
such that $\bm{Z} $ belongs to the horizontal space $\mathbb{H}_{\bm{\Phi}}$, i.e.,  $\bm{Z}^T \bm{\Phi} = \bm{0}$. Here, the matrix $\bm{F}$ has orthogonal rows and satisfies the conditions  $\tilde{\bm{\Phi}}\bm{F}^T=\bm{0}$ and $\bm{F}\bm{F}^T=\bm{I}$, where $\tilde{\bm{\Phi}}=\mathrm{diag}\{ \bm{\phi}_{1}^T,\bm{\phi}_{2}^T,\cdots,\bm{\phi}_{r}^T\}$ is a $r\times nr$ matrix with orthogonal rows. 

\vspace{6pt}
To prove Proposition 3.1, note that, 
\begin{equation} \label{eq:y_to_Z_b}
\tilde{\bm{\Phi}}\text{vec}(\bm{Z})=0
\end{equation}
when $\bm{Z}$ belongs to the horizontal space of $\bm{\Phi}$, i.e., $\bm{Z}^T \bm{\Phi} = \bm{0}$. 
Then, if we can choose a $(nr-r)\times nr$ matrix $\bm{F}$ which has orthogonal rows and satisfies $\tilde{\bm{\Phi}}\bm{F}^T=\bm{0}$, we have
\begin{equation} 
\begin{pmatrix}
\tilde{\bm{\Phi}}  \\ 
\bm{F}
\end{pmatrix} \text{vec}(\bm{Z}) = \begin{pmatrix}
\bm{0} \\ 
\bm{y}
\end{pmatrix}
\end{equation}
where $\bm{y}$ is a $(nr-r)$-dimensional column vector. 
Because $\bm{F}\text{vec}(\bm{Z}) = \bm{y}$ is an underdetermined system, one can find $\text{vec}(\bm{Z})$ by $\bm{F}^{\dagger}\bm{y}$ where $\bm{F}^{\dagger}=\bm{F}^T(\bm{F}\bm{F}^T)^{-1}$. Next, if we further impose the condition $\bm{F}\bm{F}^T = \bm{I}$, then, $\bm{F}^{\dagger}=\bm{F}^T$ and $\text{vec}(\bm{Z})=\bm{F}^{T}\bm{y}$ as shown in Proposition 3.1. 
It is straightforward to verify that, when $\text{vec}(\bm{Z})=\bm{F}^{T}\bm{y}$, we have $ \tilde{\bm{\Phi}}\text{vec}(\bm{Z})=\tilde{\bm{\Phi}}\bm{F}^{T}\bm{y}=\bm{0}$, which implies that $\bm{Z}$ belongs to the horizontal space of $\bm{\Phi}$ according to (\ref{eq:y_to_Z_b}). 

\vspace{6pt}
Proposition 3.1 implies that, from a vector $\bm{y}\in\mathbb{R}^{nr-r}$, it is possible to compute a matrix $\bm{Z}=\mathrm{Mat}_{n, r}(\bm{F}^{\dagger}\bm{y})$ that comes from the horizontal space $\mathbb{H}_{\bm{\Phi}}$. In fact, it is not hard to understand why the dimension of $\bm{y}$ is $nr-r$. Although there are $n\times r$ elements in the matrix $\bm{Z}$, the constraint $\bm{Z}^T \bm{\Phi} = \bm{0}$ consists of a number of $r$ linear equations (i.e., $\bm{Z}$ must come from the horizontal space of $\bm{\Phi}$) takes away $r$ degree of freedom, leaving behind $nr-r$ degrees of freedom. 

It is also noted that, the choice of $\bm{F}$ is obviously not unique as long as the matrix has orthogonal columns and satisfies $\tilde{\bm{\Phi}}\bm{F}^T=\bm{0}$ and $\bm{F}\bm{F}^T=\bm{I}$. In our implementation, a computationally efficient procedure is used to construct the matrix $\bm{F}$. We let $\tilde{\bm{\Phi}}=[\tilde{\bm{\Phi}}_1,\tilde{\bm{\Phi}}_2,\cdots,\tilde{\bm{\Phi}}_r]$, and let $\tilde{\bm{F}}=[\tilde{\bm{F}}_1,\tilde{\bm{F}}_2,\cdots,\tilde{\bm{F}}_r]$ be a $(nr-r)\times nr$ \textit{proposal} matrix where each block $\tilde{\bm{F}}_i$ contains non-zero entries only from row $(i-1) \times (n-1)+1$ to row $i \times (n-1)$. Then, the Gram-Schmidt (G-S) process can be applied, in parallel, to the row vectors of each block $\bigl(\begin{smallmatrix}
\tilde{\bm{\Phi}}_i\\ 
\tilde{\bm{F}}_i
\end{smallmatrix}\bigr)$ for $i=1,2,\cdots,r$. After the G-S process, each non-zero row is normalized, yielding $\bigl(\begin{smallmatrix}
\tilde{\bm{\Phi}}_i\\ 
\bm{F}_i
\end{smallmatrix}\bigr)$ and $\bm{F}=[\bm{F}_1,\bm{F}_2,\cdots,\bm{F}_r]$ that satisfies the conditions needed, i.e., orthogonal rows, $\tilde{\bm{\Phi}}\bm{F}^T=\bm{0}$ and $\bm{F}\bm{F}^T=\bm{I}$.

\vspace{6pt}
Following Proposition 3.1,  for any basepoint $p\in \mathcal{G}(r,n)$ and the tangent space $\mathcal{T}_p\mathcal{G}(r,n)$, we can define the following mapping from $\mathbb{R}^{nr-r}$ to $\mathcal{G}(r,n)$,
\begin{equation}
\mathcal{P}_p:  \bm{y}\in\mathbb{R}^{nr-r} \mapsto \pi(\bm{\Phi} \bm{V}_{\bm{Z}}\cos(\bm{\Sigma}_{\bm{Z}}) + \bm{U}_{\bm{Z}}\sin(\bm{\Sigma}_{\bm{Z}}))\in\mathcal{G}(r,n). 
\label{eq:mapping}
\end{equation}
Here, $\mathcal{P}_p := \tilde{\mathcal{P}}_p \circ \text{Exp}_p$ where $\tilde{\mathcal{P}}_p$ maps an $(nr-r)$-dimensional vector to the tangent space $\mathcal{T}_p\mathcal{G}(r,n)$, 
\begin{equation}
\tilde{\mathcal{P}}_p:  \bm{y}\in\mathbb{R}^{nr-r} \mapsto v \in \mathcal{T}_p\mathcal{G}(r,n)
\end{equation}
and the Exponential map, $\text{Exp}_p$, is defined in (\ref{eq:ExpMap}).

\subsection{Injectivity of the map, $\mathcal{P}_p$} Although $\mathcal{P}_p$ in (\ref{eq:mapping}) defines a map from $\mathbb{R}^{nr-r}$ to $\mathcal{G}(r,n)$, it is important to guarantee that such a map is injective. Proposition 3.2 below gives the condition for which $\mathcal{P}_p$ is injective. 

\vspace{6pt}
Proposition 3.2.
\label{prop:2}
Given a basepoint $p \in \mathcal{G}(r,n)$, there exists an open set $\Omega_p \subset \mathbb{R}^{nr-r}$
\begin{equation}
\Omega_p:=\left\{\bm{y} \in  \mathbb{R}^{nr-r}; ||\bm{y}|| < \frac{\pi}{2} \right\}
\label{eq:Omega_set}
\end{equation}
such that the mapping $\mathcal{P}_p$ is injective. 

\vspace{6pt}
To prove Proposition 3.2, consider an open set $D_p \subset \mathcal{T}_p\mathcal{G}(r,n)$
\begin{equation} \label{eq:v_radius}
D_p:=\left\{v \in \mathcal{T}_p\mathcal{G}(r,n); ||v|| < \pi/2 \right\}
\end{equation}
where $\pi/2$ is the injective radius on the Grassmann manifold \citep{Friderikos2022}. Hence, we seek an open set $\Omega_p \subset \mathbb{R}^{nr-r}$ such that $\mathcal{A}_p$ maps a vector $\bm{y} \in \Omega_p$ to the open disk $D_p$ above. Because $\bm{Z} = \text{Mat}_{n,r}(\bm{F}^{\dagger}\bm{y})$, we have
\begin{equation}
\begin{split}
\bm{Z}^T\bm{Z} & = \begin{bmatrix}
\bm{y}^T \bm{F}_1 \\ \bm{y}^T \bm{F}_2 \\  \vdots \\  \bm{y}^T \bm{F}_r
\end{bmatrix}
\begin{bmatrix}
\bm{F}_1^T\bm{y} & \bm{F}_2^T\bm{y}  & \cdots & \bm{F}_r^T\bm{y} 
\end{bmatrix} \\
& = \begin{bmatrix}
\sum_{i=1}^{n-1}y_{(i-1)\times r + 1}^2 &  &  & \\ 
 & \sum_{i=1}^{n-1}y_{(i-1)\times r + 2}^2 &  & \\ 
 &  & \ddots  & \\ 
 &  &  & \sum_{i=1}^{n-1}y_{(i-1)\times r + r}^2
\end{bmatrix}
\end{split}
\end{equation}

Because the Riemannian metric on $\mathcal{G}(r,n)$ is defined by $\left \langle v_1,v_2 \right \rangle_p = \text{trace}(\bm{Z}_1^T\bm{Z}_2)$ given the basepoint $p$, we have
\begin{equation}
||v|| = \sqrt{\text{trace}(\bm{Z}^T\bm{Z})} = \left(\sum_{i=1}^{nr-r}y_{i}^2\right)^{\frac{1}{2}} = ||\bm{y}||
\end{equation}
which needs to be smaller than $\pi/2$ according to (\ref{eq:v_radius}). 
Hence, we have an open set $\Omega_p$, defined in (\ref{eq:Omega_set}), such that $\mathcal{P}_p$ is injective.

\subsection{Prediction using the pGP regression}
\begin{figure}[h!] 
\includegraphics[width=\textwidth]{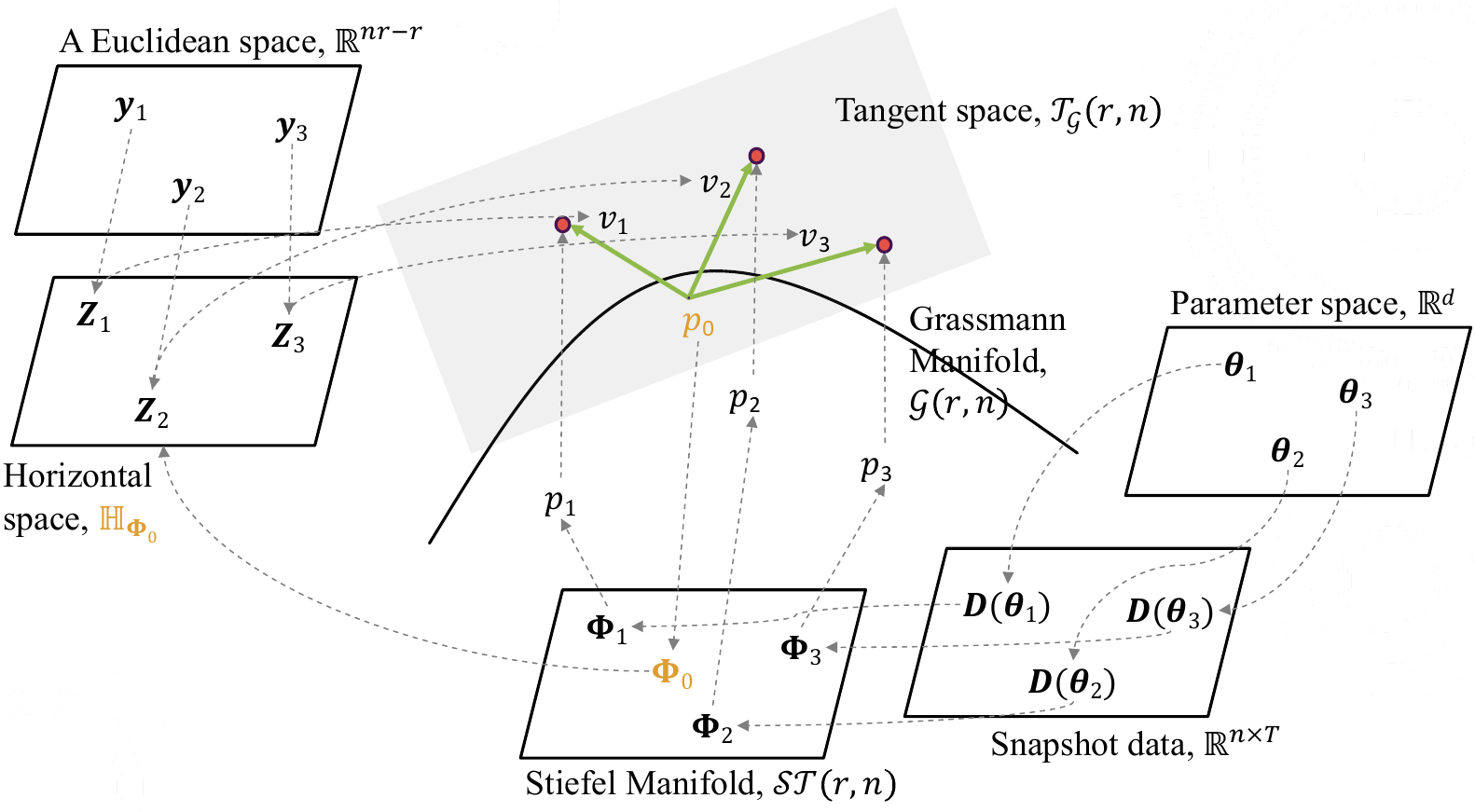}
\caption{A sketch of the key relationships between different spaces/manifolds involved in the pGP regression.}
\label{fig:framework}
\end{figure}

Propositions 3.1 and 3.2 suggest an effective approach to establish the mapping between the parameter and POD basis. In a nutshell, as illustrated in Figure \ref{fig:framework}, a GP can be established to describe the mapping between the parameters space $\mathbb{R}^{d}$ and $\mathbb{R}^{nr-r}$, and then, the GP is projected to the Grassmann manifold. As a result, for any new parameter $\bm{\theta}^*$, a vector $\bm{y}^*$ can be found and mapped back to the Grassmann manifold to determine the optimal subspace onto which the full-order system is projected (equivalently, the POD basis matrix $\bm{\Phi}^*$). 

Let the vector $\bm{y}$ follow a multivariate Gaussian distribution, i.e., $\bm{y}\sim \mathcal{N}(\bm{0}_{nr-r,1}, \bm{K})$ where $\bm{K}$ is a $(nr-r)\times (nr-r)$ symmetric and positive semi-definite matrix. Then, $\bm{Z}=\mathrm{Mat}_{n, r}(\bm{F}^{\dagger}\bm{y})$ follows a matrix-variate Gaussian, $\bm{Z}\sim \mathcal{N}(\bm{0}_{n \times r}, \bm{F}^{\dagger} \bm{K} (\bm{F}^{\dagger} )^T)$ in the sense that $\text{vec}(\bm{Z})\sim \mathcal{N}(\bm{0}_{nr\times 1}, \bm{F}^{\dagger} \bm{K} (\bm{F}^{\dagger} )^T)$. 
For a given point $\mu\in\mathcal{G}(r,n)$ and a symmetric positive semi-definite matrix $\bm{K}\in\mathbb{R}^{(nr-r)\times (nr-r)}$, we say that a random point $X\in\mathcal{G}(r,n)$ follows a projected Gaussian Distribution (pGD) on $\mathcal{G}(r,n)$
\begin{equation}
X \sim \mathcal{P}_{\mu}(\mathcal{N}(\bm{0}_{nr-r,1},\bm{K}))
\end{equation}
where the projection, $\mathcal{P}_{\mu}: \mathbb{R}^{nr-r} \rightarrow \mathcal{G}(r,n)$, is associated with the given point $\mu$. We denote the pGD by $\mathcal{N}_{\mathcal{G}}(\mu, \bm{K})$, 
and write $X \sim \mathcal{N}_{\mathcal{G}}(\mu, \bm{K})$, $\mu := \mu_{\mathcal{G}}(X)\in\mathcal{G}(r,n)$, and $\text{cov}_{\mathcal{G}}(X):=\bm{K}$. 


To obtain the mapping $\mathcal{P}_{\mu}$ that connects $\mathbb{R}^{nr-r}$ and $\mathcal{G}(r,n)$ given a point $\mu\in \mathcal{G}(r,n)$ and $\bm{\Phi} =\pi^{-1}(\mu) \in \mathcal{ST}(r,n)$, we first introduce an operator $\mathcal{P}_1$ such that $\mathcal{P}_1(\bm{y})=\mathrm{Mat}_{n, r}(\bm{F}^{\dagger}\bm{y})=\bm{Z}$. Then, let $\bm{Z} = \bm{U}_{\bm{Z}}\bm{\Sigma}_{\bm{Z}} \bm{V}_{\bm{Z}}^T$ be the thin SVD, an operator $\mathcal{P}_2$ can be introduced that operates on the SVD of $\bm{Z}$ such that $\mathcal{P}_2(\bm{Z}) = \bm{\Phi} \bm{V}_{\bm{Z}}\cos(\bm{\Sigma}_{\bm{Z}}) + \bm{U}_{\bm{Z}}\sin(\bm{\Sigma}_{\bm{Z}})$ yields a $n\times r$ matrix that spans a subspace in $\mathcal{G}(r,n)$. 
Hence,  the mapping $\mathcal{P}_{\mu}$ can be defined as 
\begin{equation}
\mathcal{P}_{\mu} := \text{span}\circ \mathcal{P}_2\circ \mathcal{P}_1
\end{equation} 
that maps a $(nr-r)\times 1$ vector from $\mathbb{R}^{nr-r}$ to a $r$-dimensional subspace in $\mathcal{G}(r,n)$. \\


\vspace{6pt}
Proposition 3.3.
\label{prop:3}
    If $X_i \sim \mathcal{N}_{\mathcal{G}_i}(\mu_i, \bm{K}_{i})$ for $i=1,2$, then, $(X_1, X_2)$ are jointly pGD on $\mathcal{G}_1 \times \mathcal{G}_2$
    \begin{equation}
(X_1, X_2) \sim \mathcal{N}_{\mathcal{G}_1 \times \mathcal{G}_2}
 \left ( \begin{pmatrix}
\mu_1\\ 
\mu_2
\end{pmatrix} ,
\tilde{\bm{K}}\right ),
\quad\quad
\tilde{\bm{K}}= \begin{pmatrix}
\bm{K}_{1}, \bm{K}_{12}\\ 
\bm{K}_{12}^T, \bm{K}_2
\end{pmatrix} 
\end{equation}
Then, the conditional distribution of $X_1$ given $X_2=p_2$ is given by
\begin{equation} \label{eq:conditional}
X_1 \mid (X_2=p_2) \sim \mathcal{P}_{\mu_1}(\mathcal{MN}(\bm{u}_{\bm{s}}, \bm{K}_{\bm{s}}))
\end{equation}
where $\bm{u}_{\bm{s}}=\bm{K}_{12} \bm{K}_{2}^{-1}\bm{s}$,  $\bm{K}_{\bm{s}}=\bm{K}_{1}-\bm{K}_{12}\bm{K}_{2}^{-1}\bm{K}_{12}^T$, $\bm{s} = \mathcal{P}_{\mu_2}^{-1}(p_2)$, and $\mathcal{P}_{\mu_2}^{-1}$ is the inverse map.  

\vspace{6pt}
Proposition 3.3 is obtained following a similar technique in  \cite{Mallasto2018}. Let $\mathbb{A}_1=\mathcal{P}_{\mu_1}^{-1}(p_1) = \{\bm{y}\in\mathbb{R}^{nr-r}: \mathcal{P}_{\mu_1}(\bm{Y})=p_1\}$ be the preimage of $p_1$ in $\mathbb{R}^{nr-r}$ associated with $\bm{\Phi}_{\mu_1}$ given a point $\mu_1$, and let $\mathbb{A}_2=\mathcal{P}_{\mu_2}^{-1}(p_2)$ similarly be the preimage of $p_2$, then, 
\begin{equation}
\begin{split}
\text{Pr}(X_1|X_2=p_2) & = \frac{\text{Pr}(\bm{y}_1\in \mathbb{A}_1, \bm{y}_2\in \mathbb{A}_2)}{\text{Pr}(\bm{y}_2\in \mathbb{A}_2)} \\
& = \sum_{\bm{y}_1\in \mathbb{A}_1, \bm{y}_2\in \mathbb{A}_2} \frac{\mathcal{MN}(\bm{y}_2; \bm{0}, \bm{K}_{2})}{\text{Pr}(\mathbb{A}_2)} \frac{\mathcal{MN}((\bm{y}_1^T,\bm{y}_2^T)^T; \bm{0},\tilde{\bm{K}})}{\mathcal{MN}(\bm{y}_2; \bm{0}, \bm{K}_{2})} \\
& = \sum_{\bm{y}_1\in \mathbb{A}_1, \bm{y}_2\in \mathbb{A}_2} \lambda_{\bm{y}_2} \mathcal{MN}(\bm{y}_1; \bm{u}_{\bm{y}_2}, \bm{K}_{\bm{y}_2}) 
\end{split}
\end{equation}
where  $\bm{u}_{\bm{y}_2}=\bm{K}_{12} \bm{K}_{2}^{-1}\bm{y}_2$, and $\bm{K}_{\bm{y}_2}=\bm{K}_{1}-\bm{K}_{12}\bm{K}_{2}^{-1}\bm{K}_{12}^T$. 

Hence, 
\begin{equation}
X_1 \mid (X_2=p_2) \sim \mathcal{P}_{\mu_1}(\mathcal{MN}(\bm{u}_{\bm{s}}, \bm{K}_{\bm{s}}))
\end{equation}
where $\bm{s} = \mathcal{P}_{\mu_2}^{-1}(p_2)$.



\vspace{6pt}
Once a pGD is obtained as described above, a pGP on Grassmann manifold can be defined. We start with a multivariate GP defined in the Euclidean space. 
A collection $\bm{y}$ of random vectors is a Multivariate GP (MGP) in $\mathbb{R}^d$, denoted by $\bm{y} \sim \mathcal{MGP}(f, \omega, \bm{K})$, with a matrix-valued mean function $f: \mathbb{R}^d \rightarrow \mathbb{R}^{nr-r}$, kernel $\omega: \mathbb{R}^d \times \mathbb{R}^d \rightarrow \mathbb{R}$ and $\bm{K} \in \mathbb{R}^{(nr-r)\times (nr-r)}$, if any finite collection of $\bm{y}_1$, $\bm{y}_2$, $\cdots$, $\bm{y}_k$ have a joint multivariate Gaussian distribution such that 
\begin{equation}
[\bm{y}_1^T, \bm{y}_2^T, \cdot, \bm{y}_k^T]^T \sim \mathcal{N}([f^T(\bm{\theta}_1), f^T(\bm{\theta}_2), \cdot, f^T(\bm{\theta}_k)^T]^T, \bm{\Omega} \otimes \bm{K})
\end{equation}
where $\bm{\Omega} \in \mathbb{R}^{k\times k}$ with $\bm{\Omega}_{i,j}=\omega(\bm{\theta}_i, \bm{\theta}_j;\bm{\xi})$. 
Then, a pGP on $\mathcal{G}(r,n)$ can be defined. 

\vspace{6pt}
\textbf{Projected Gaussian Process.} A pGP on $\mathcal{G}(r,n)$ is defined by
\begin{equation}
g \sim \mathcal{PGP}(m, \omega, \bm{K})
\end{equation}
where a collection $(g(\bm{\theta}_1), g(\bm{\theta}_2), \cdots, g(\bm{\theta}_k))\sim \mathcal{P}_{m}(\mathcal{MGP}(\bm{0},\omega, \bm{K})$ (i.e., a joint pGD), $m(\bm{\theta}):=\mu_{\mathcal{G}}(g(\theta))$ is the basepoint function and $k(\bm{\theta}_i, \bm{\theta}_i):=\text{cov}_{\mathcal{G}}(g(\bm{\theta}_i), g(\bm{\theta}_j))=\omega(\bm{\theta}_i, \bm{\theta}_j)\bm{K}$ is the co-variance function in $\mathbb{R}^{nr-r}$. We denote the process by $g \sim \mathcal{PGP}(m, \omega, \bm{K})$. 

Following Proposition 3.3, the joint distribution between the training output $\bm{p}^{\text{train}}=(p_1,\cdots,p_k)^T$ and test output $p_*$ is given by:
\begin{equation}
\begin{pmatrix}
p_*\\ 
\bm{p}^{\text{train}}
\end{pmatrix} \sim \mathcal{N}_{\mathcal{G}_* \times \mathcal{G}_1 \times \cdots \times \mathcal{G}_k}
 \left ( \begin{pmatrix}
m_*\\ 
\bm{m}^{\text{train}}
\end{pmatrix} ,
 \begin{pmatrix}
\bm{K}_{**}, \bm{K}_{*}\\ 
\bm{K}_{*}^T, \bm{K}^{\text{train}}
\end{pmatrix} \right )
\end{equation}
where $m_*=m(\theta_*)$, $\bm{m}^{\text{train}}=(m(\bm{\theta}_1),m(\bm{\theta}_2),\cdots,m(\bm{\theta}_d))^T$, $\bm{K}^{\text{train}}= \bm{\Omega} \otimes \bm{K} \in \mathbb{R}^{k(nr-r)\times k(nr-r)}$, $K_{**}=\omega(\bm{\theta}_*,\bm{\theta}_*)\bm{K} \in \mathbb{R}^{(nr-r)\times (nr-r)}$, 
$\bm{K}_*=(\omega(\bm{\theta}_*,\bm{\theta}_1),\omega(\bm{\theta}_*,\bm{\theta}_2),\cdots, \omega(\bm{\theta}_*,\bm{\theta}_d)) \otimes \bm{K} \in \mathbb{R}^{(nr-r)\times k(nr-r)}$. 

Finally, 
\begin{equation} \label{eq:pred_mean}
p_*|\bm{p}^{\text{train}} \sim \mathcal{P}_{m_*}(\mathcal{MN}(\bm{u}_*,\tilde{\bm{K}}_{*}))
\end{equation}
where 
\begin{equation}  \label{eq:pred_mean_u}
\bm{u}_{*}=\bm{K}_{*}(\bm{K}^{\text{train}})^{-1}( (\mathcal{P}_{m(\bm{\theta}_1)}^{-1}(p_1))^T, (\mathcal{P}_{m(\bm{\theta}_2)}^{-1}(p_2))^T,\cdots,(\mathcal{P}_{m(\bm{\theta}_d)}^{-1}(p_k))^T)^T
\end{equation}
and
\begin{equation}  \label{eq:pred_mean_K}
\tilde{\bm{K}}_{*}=\bm{K}_{**}-\bm{K}_{*}(\bm{K}^{\text{train}})^{-1}\bm{K}_{*}^T.
\end{equation}

Hence, a point estimate of the subspace $p_* \in \mathcal{G}(r,n)$ can be obtained by $\hat{p}_* = \mathcal{P}_{m_*}(\bm{u}_*)$. Because $\bm{u}_*$ is the mean of the multivariate Gaussian distribution $\mathcal{MN}(\bm{u}_*,\tilde{\bm{K}}_{*})$ defined in $\mathcal{R}^{nr-r}$, $\hat{p}_*$ is a maximum a posteriori probability (MAP) estimate. Also recall that $||\bm{u}_*||$ needs to be smaller than $\pi/2$ for $\mathcal{P}_{m_*}$ to be injective. Hence, if $||\bm{u}_*||>\pi/2$, we shrink this vector $\bm{u}_* \leftarrow \frac{\pi}{2}\frac{\bm{u}_*}{||\bm{u}_*||}$, which is the projection of the original vector  $\bm{u}_*$ onto the surface of a $(nr-r)$-dimensional sphere.
Similarly, the estimate $\hat{\bm{\Phi}}_*$ is obtained as $\mathcal{P}_2\circ \mathcal{P}_1(\bm{u}_*)$ such that $\hat{p}_*$ is spanned by $\hat{\bm{\Phi}}_*$. 

\begin{figure}[h!] 
\centering
\includegraphics[width=0.65\textwidth]{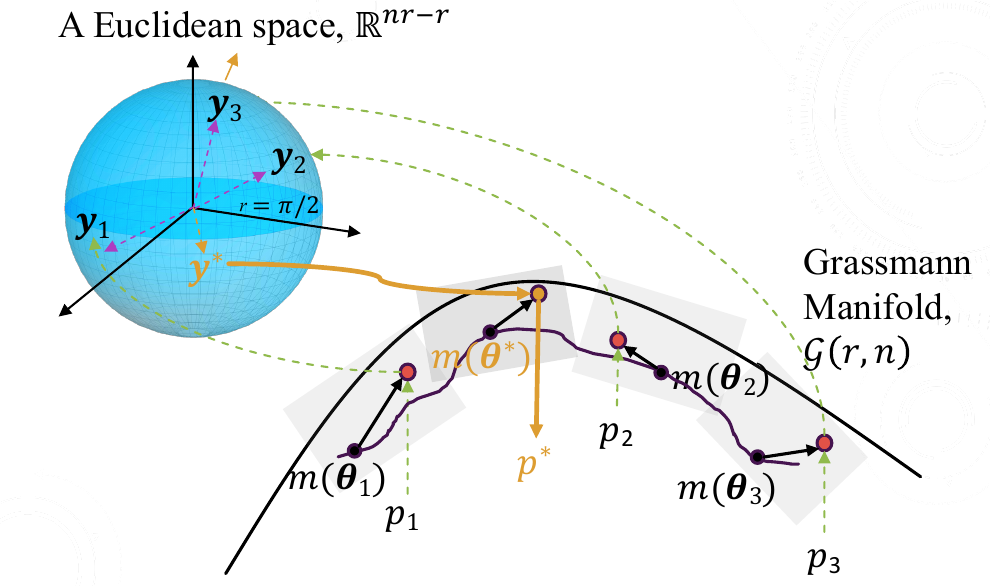}
\caption{An illustration of the subspace prediction using the pGP regression.}
\label{fig:pGP}
\end{figure}

Figure \ref{fig:pGP} illustrates the process of predicting the optimal subspace using pGP. In this illustration, suppose that we obtain the snapshot data from three parameter settings $\bm{\theta}_1$, $\bm{\theta}_2$ and $\bm{\theta}_3$, as well as the optimal subspaces $p_1$, $p_2$ and $p_3$ spanned by the POD modes found at $\bm{\theta}_1$, $\bm{\theta}_2$ and $\bm{\theta}_3$. The prediction of the optimal subspace involves the following steps:

(i) Given the mean function $m(\bm{\theta})$, the three subspaces $p_1$, $p_2$ and $p_3$ are firstly mapped to the Euclidean space $\mathbb{R}^{nr-r}$ through the projection $\mathcal{P}^{-1}_{m(\bm{\theta})}$; see (\ref{eq:pred_mean_u}). Note that, $(\mathcal{P}_{m(\bm{\theta})}^{-1}(p))^T$ returns a $(nr-r)\times 1$ column vector on $\mathbb{R}^{nr-r}$. 

(ii) The three column vectors $(\mathcal{P}_{m(\bm{\theta}_1)}^{-1}(p_1))^T$, $(\mathcal{P}_{m(\bm{\theta}_2)}^{-1}(p_2))^T$ and $(\mathcal{P}_{m(\bm{\theta}_3)}^{-1}(p_3))^T$ are used to predict a new column vector $\bm{u}_{*}$, for the new parameter $\bm{\theta}_*$, through a linear combination defined by a GP; see (\ref{eq:pred_mean_u}). 

(iii) The covariance matrix $\tilde{\bm{K}}_{*}$ is computed, which is exactly the same as the convectional GP. Steps (ii) and (iii) together yield a multivariate Gaussian distribution of $\bm{y}_{*}$ on $\mathbb{R}^{nr-r}$, i.e., $\bm{y}_{*}\sim \mathcal{MN}(\bm{u}_*,\tilde{\bm{K}}_{*})$. 

(iv) Finally, the Gaussian distribution of $\bm{y}_{*}$ on $\mathbb{R}^{nr-r}$ (or, the predicted column vector $\bm{u}_{*}$ if only a point prediction is needed) is projected back to the Grassmann manifold through the projector $\mathcal{P}_{m(\bm{\theta}_*)}$.

\subsection{Discussions on model specification and parameter estimation} \label{sec:discussions}
A close examination of (\ref{eq:pred_mean}) reveals an important \textit{difference} between how predictions are made by conventional GP and the proposed pGP. Using a conventional GP, the point predictor is a linear combination of the observations and the weights are determined by the covariance function (i.e., the point prediction does not depend on the mean of the GP; a property well-known in the literature). For the proposed pGP, as indicated by (\ref{eq:pred_mean}), the point prediction does depend on the function $m(\bm{\theta})$ that describes the relationship between the parameter and basepoint on the Grassmann manifold. In other words, the choice of $m(\bm{\theta})$ determines the mapping $\mathcal{P}_{m}$. Two approaches can be used to determine the function $m(\bm{\theta})$. The first approach is to fix the basepoint for all parameters, i.e., $m(\bm{\theta})=\mu$. which can be obtained using a global POD basis. The second approach is to use another regression model to determine $m(\bm{\theta})$, such as the tree-based method described in  \cite{Liu2023}. 

In addition, because $n$ is usually large, a practical way to parameterize the $(nr-r)\times (nr-r)$ covariance matrix $\bm{K}$ is to let $\bm{K}=\sigma_{\bm{K}} \bm{I}$ and estimate $\sigma_{\bm{K}}$ from data. As for the kernel function $\omega(\bm{\theta}_i,\bm{\theta}_j; \bm{\xi})$, existing Gaussian, Exponential or other kernel functions can be used, and the parameter $\bm{\xi}$ can be estimated from data. Hence, once $m(\bm{\theta})$ is determined,  $(\sigma_{\bm{K}}, \bm{\xi})$ can be estimated using existing methods such as the cross-validation-based approach or maximum likelihood estimation described in \cite{Rasmussen2005}.

Also note that, for systems with high-dimensional parameter spaces (i.e., relatively large $d$), it is often the case that not all parameters are equally important in terms of their predictive power of the POD modes. In other words, the POD modes can be sensitive to some parameters while independent of the others. Hence, when constructing the statistical learning model that establishes the mapping from parameter space to Grassmann manifold, it is useful to identify those relevant parameters. 
To illustrate the idea, consider the squared Exponential kernel given by
\begin{equation}
\omega(\bm{\theta}_i,\bm{\theta}_j) =\xi_1^2 \delta_{\{\bm{\theta}_i=\bm{\theta}_j\}}+  \xi_2^2 \exp \left\{-\frac{(\bm{\theta}_i-\bm{\theta}_j)^T \bm{K}_{\omega} (\bm{\theta}_i-\bm{\theta}_j)}{2} \right\} 
\end{equation}
where $\bm{K}_{\omega} = \text{diag}\{\xi_3^{-2},\xi_4^{-2},\cdots,\xi_{d+2}^{-2}\}$. Here, the parameters $\xi_3, \xi_4, \cdots, \xi_{d+2}$ play the role of characteristic length-scales for parameter $\bm{\theta}_1, \bm{\theta}_2, \cdots, \bm{\theta}_d$. If the estimated length-scales are very large, the corresponding parameters have little effects on the prediction and are practically removed from the model. This is known as the Automatic Relevance Determination (ARD) \citep{Rasmussen2005} and more details can be found in Sec.\ref{sec:numerical}. 

Finally, it is worth noting that the proposed pGP naturally involves the use of the global POD basis as its special case. Take the squared Exponential kernel above as an example.
If $\xi_1=0$ and the estimated $\xi_3, \xi_4, \cdots, \xi_{d+2}$ are all close to zero, $\omega(\bm{\theta}_i,\bm{\theta}_j)$ becomes practically zero. In this case, (\ref{eq:pred_mean_u}) immediately implies that $\bm{u}_*=\bm{0}$ and $p_*|\bm{p}^{\text{train}} \sim \mathcal{P}_{m_*}(\mathcal{MN}(\bm{0},\tilde{\bm{K}}_{*}))$. If the global POD basis is chosen as the reference point (which spans a subspace in the Grassmann manifold where the tangent space is attached to) and the basepoint function $\bm{m}$ is invariant with $\bm{\theta}$, then, any subspace $p_i$, corresponding to the parameter $\bm{\theta}_i$, on the Grassmann manifold can be seen as \textit{i.i.d.} samples from a pGD, $\mathcal{P}_{m_*}(\mathcal{MN}(\bm{0},\tilde{\bm{K}}))$. Hence, the point prediction, $\hat{p}_*$, is just the subspace spanned by the global POD basis. More details can be found in Sec.\ref{sec:numerical}.

\section{Numerical Examples} \label{sec:numerical}

Numerical investigations are performed to demonstrate the performance of the proposed pGP approach. In particular, we compare the prediction accuracy between the proposed method and the existing ones, illustrate the uncertainty quantification capabilities and how the pGP can help to identify important parameters. 


\textbf{Experiment setup and data.} 
Consider, $\mathcal{A}({\bm{\theta}})x(t,\bm{s}) = 0$, an advection-diffusion process,  
where $x(t,\bm{s})$ is a spatio-temporal process in space $\bm{s} \in \mathbb{S} \subset  \mathbb{R}^2$ and time $t \in [0,T]$, and $\mathcal{A}({\bm{\theta}})$ is the advection-diffusion operator parameterized by $\bm{\theta}$:
\begin{equation} \label{eq:A}
\mathcal{A}x := \dot{x} + \bm{\vec{v}}^T\triangledown x - \triangledown \cdot [\bm{D} \triangledown  x]
\end{equation}	
where $\bm{\vec{v}}$, $\bm{D}$, $\triangledown$ and $\triangledown\cdot$ represent the velocity, diffusivity, decay, gradient and divergence, respectively. In this experiment, $\mathbb{S}$ is a 2D rectangular domain.

To investigate the performance of adapting the POD basis using the proposed pGP, we first let $\bm{v} = (1, 0)^T$ and $\bm{D} =\mathrm{diag}\{d_1,d_2\}$, and generate data by letting $(d_1,d_2)$ take values from a 2D mesh grid with uniformly spaced $x$- and $y$-coordinates in the interval $[0.01,0.05]$ with a step size of 0.002 (more parameters will be included in subsequent investigations). Hence, a number of $21\times 21=441$ parameter settings are considered, and the process is simulated for each parameter setting. 

\begin{figure}[h!] 
\includegraphics[width=\textwidth]{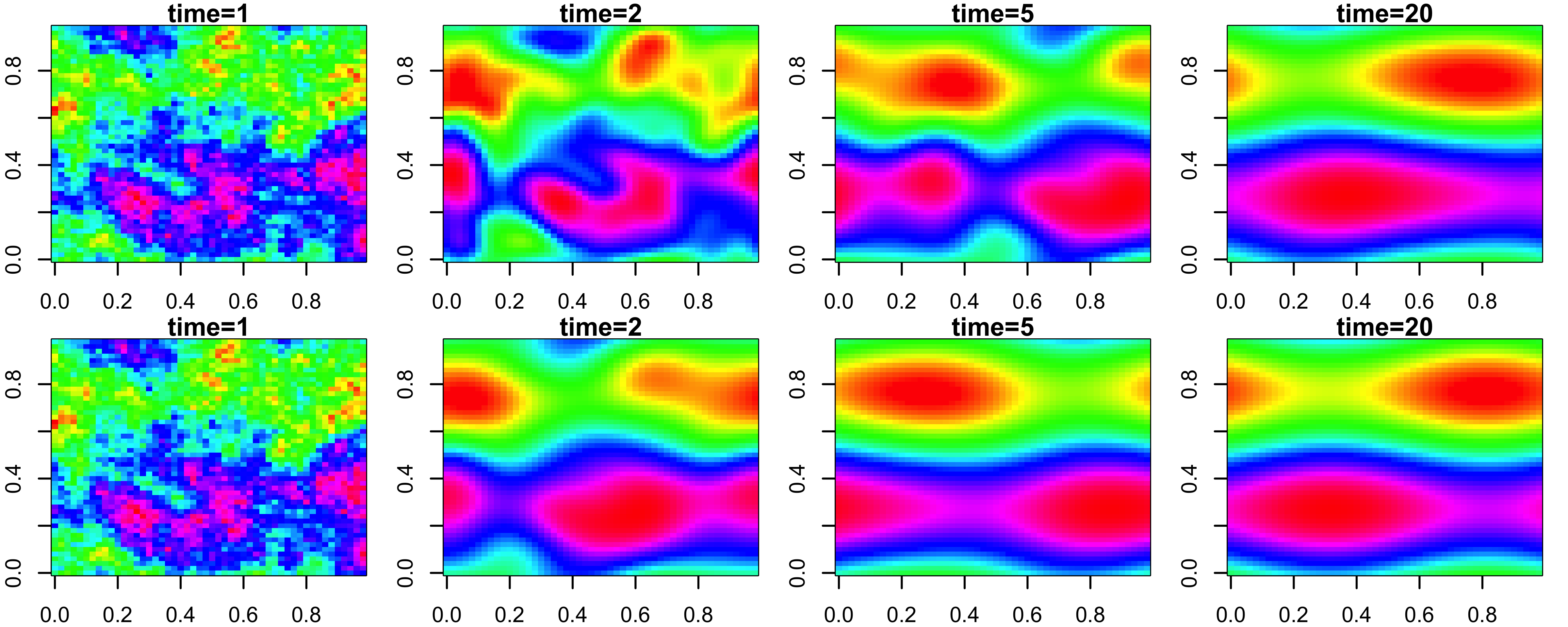}
\caption{Two simulated processes respectively at parameter settings $[0.01,0.01]$ (top) and $[0.05,0.05]$ (bottom) of $(d_1,d_2)$ .}
\label{fig:ad_data}
\end{figure}

\begin{figure}[h!] 
\includegraphics[width=\textwidth]{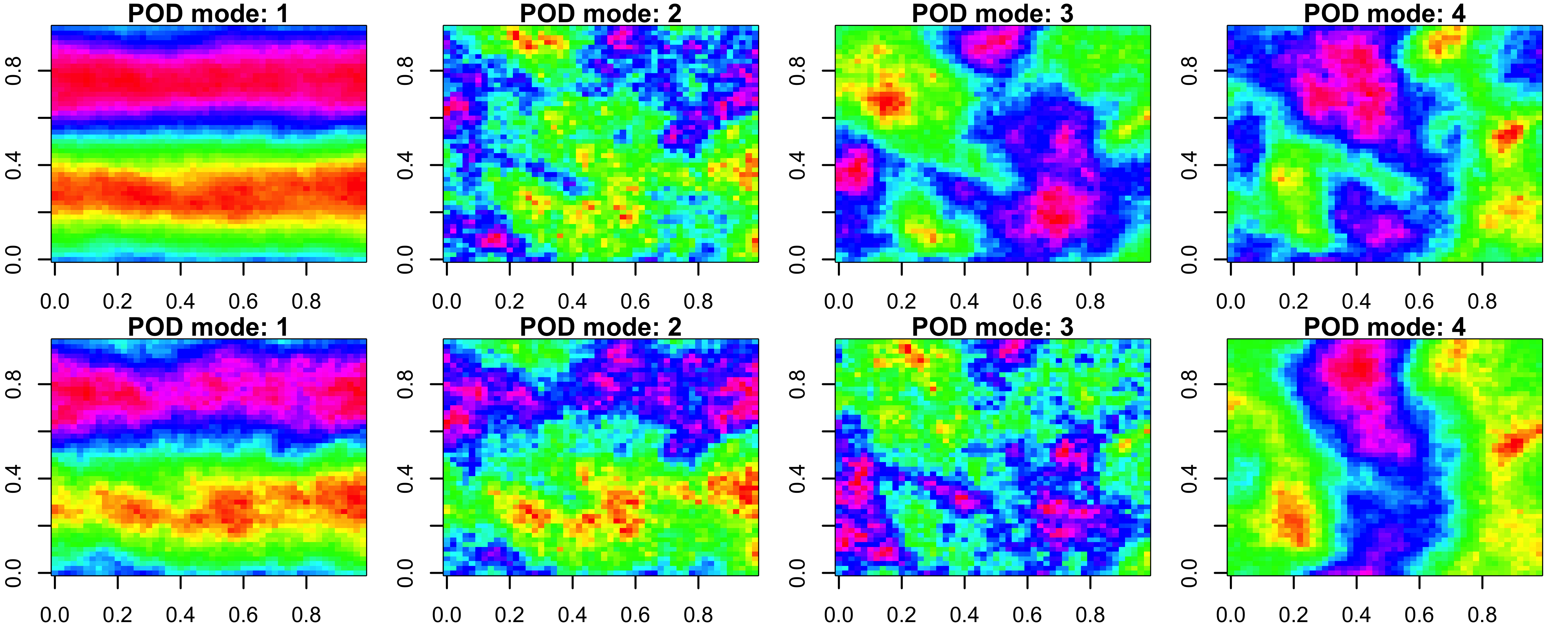}
\vspace{-18pt}
\caption{The four leading POD modes obtained from the simulated processes at parameter settings $[0.01,0.01]$ (top) and $[0.05,0.05]$ (bottom) of $(d_1,d_2)$.}
\label{fig:ad_POD}
\end{figure}

For illustrative purposes, Figure \ref{fig:ad_data} shows two simulated processes respectively for two parameter settings $[0.01,0.01]$ and $[0.05,0.05]$ of  $(d_1,d_2)$, and Figure \ref{fig:ad_POD} shows the four leading POD modes of the two processes. It is seen that the POD modes vary as the parameters $d_1$ and $d_2$ change, suggesting the potential need to adapt the POD basis for parameters change. 

\textbf{Performance metrics.}
To comprehensively evaluate the accuracy of the predicted POD bases, four performance metrics are adopted in our  investigations: 

\vspace{6pt}
$\bullet$ Recall that the optimal POD basis $\bm{\Phi}$ is found by minimizing the $\ell_2$ error, $\mathrm{min}_{\bm{\Phi}}||  \bm{D}(\bm{\theta})- \bm{\Phi}\bm{\Phi}^T \bm{D}(\bm{\theta})||^2_F$, where $\bm{D}$ is the snapshot data and $||\cdot||_F$ is the Frobenius norm of matrices in the vector space. Hence, the $\ell_2$ error can be naturally adopted to assess the accuracy of the predicted POD bases. In particular, let $\hat{\bm{\Phi}}(\bm{\theta})$ be the predicted POD basis for the parameter setting $\bm{\theta} \in \Theta^{\text{test}}$, the $\ell_2$ error at this parameter setting is given by
\begin{equation}
e_F(\bm{\theta}) = ||  \bm{D}(\bm{\theta})- \hat{\bm{\Phi}}(\bm{\theta})\hat{\bm{\Phi}}(\bm{\theta})^T \bm{D}(\bm{\theta})||^2_F.
\end{equation}

$\bullet$ Following the $\ell_2$ error above, we also report the relative error so as to assess the percentage increase of the $\ell_2$ error from the optimal to the predicted POD basis, 
\begin{equation} \label{eq:e_R}
e_R(\bm{\theta}) =  \frac{e_F(\bm{\theta}) - e^*(\bm{\theta})}{e^*(\bm{\theta})} \times 100\%
\end{equation}
where $e^*(\bm{\theta})$ is the minimized $\ell_2$ error based on the optimal POD basis obtained from solving $\mathrm{min}_{\bm{\Phi}}||  \bm{D}(\bm{\theta})- \bm{\Phi}\bm{\Phi}^T \bm{D}(\bm{\theta})||^2_F$ using data from the parameter setting $\bm{\theta} \in \Theta^{\text{test}}$. 

$\bullet$ In addition, motivated by the subspace angle interpolation of POD bases, the third performance metric is obtained by computing the geometric distance, i.e., the principal angel, between the subspaces respectively spanned by the optimal POD basis and the predicted POD basis. 
\begin{equation}
e_A(\bm{\theta}) = \left( \sum_{i=1}^{r} \text{arccos}^2(\sigma_i) \right)^{\frac{1}{2}}
\end{equation}
where $\sigma_1 \geq \sigma_2 \geq \cdots \geq \sigma_r$ are the singular values of $(\bm{\Phi}^*)^T\hat{\bm{\Phi}}$ with $\bm{\Phi}^*$ being the optimal POD basis  from solving $\mathrm{min}_{\bm{\Phi}}||  \bm{D}(\bm{\theta})- \bm{\Phi}\bm{\Phi}^T \bm{D}(\bm{\theta})||^2_F$. 

$\bullet$ Finally, we evaluate the worst case scenario by computing the $\ell_\infty$ prediction error as follows:
\begin{equation}
e_I(\bm{\theta}) = ||  (\bm{D}(\bm{\theta})- \hat{\bm{\Phi}}(\bm{\theta})\hat{\bm{\Phi}}(\bm{\theta})^T \bm{D}(\bm{\theta}))^T ||_\infty
\end{equation}
where $|| \cdot ||_\infty$ is the matrix infinity norm which shows the largest error among the columns of the matrix $\bm{D}(\bm{\theta})- \hat{\bm{\Phi}}(\bm{\theta})\hat{\bm{\Phi}}(\bm{\theta})^T \bm{D}(\bm{\theta})$. 
\vspace{6pt}

\textbf{Numerical results.} The data are divided into a training set and a testing set. The training set includes 36 parameter settings $\bm{\theta}=(d_1,d_2) \in \Theta^{\text{train}}$ from a coarse 2D mesh grid with uniformly spaced $x$- and $y$-coordinates in the interval $[0.01,0.05]$ with a larger step size of 0.008. Data generated from the remaining 405 parameter settings, $\bm{\theta} \in \Theta^{\text{test}}$, are reserved for testing. 

\begin{figure}[h!] 
\includegraphics[width=0.95\textwidth]{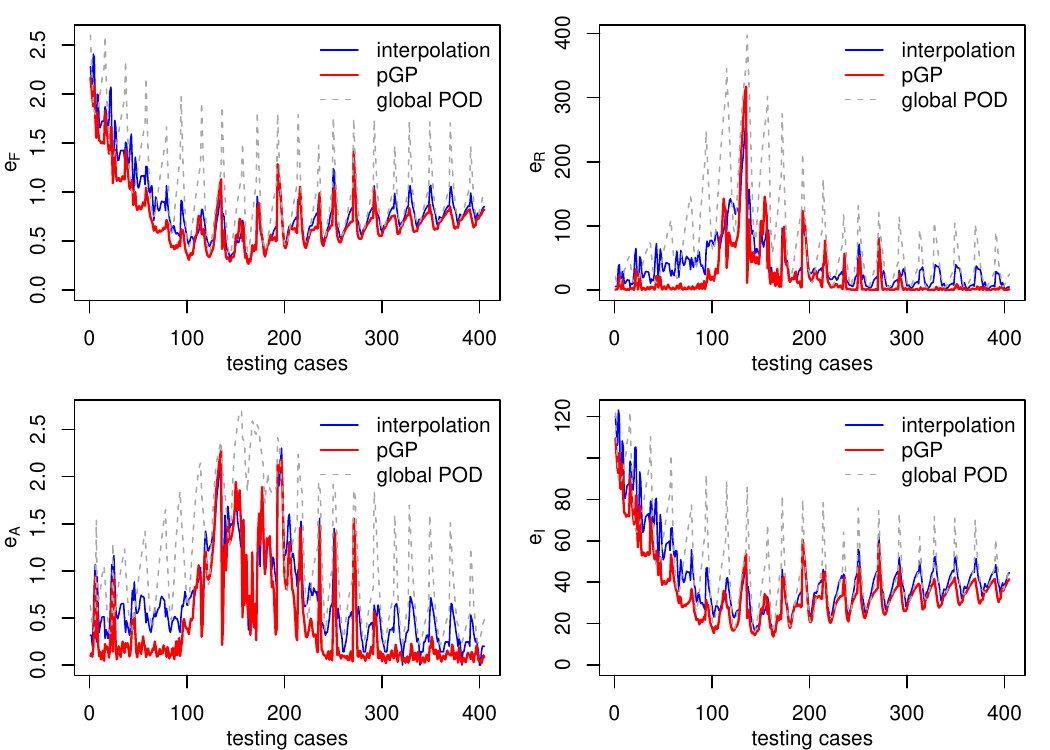}
\vspace{-8pt}
\caption{Testing errors, $e_F$, $e_R$, $e_A$ and $e_I$, for all 405 testing cases ($r=5$)}
\label{fig:error_K5}
\end{figure}

Figure \ref{fig:error_K5} shows the four types errors, $e_F$, $e_R$, $e_A$ and $e_I$, for all 405 testing cases while letting the reduced dimension to be $r=5$. Here, three different approaches are used, including the POD interpolation (using a Gaussian radial kernel), the global POD approach, and the proposed pGP approach. It is noted that, the proposed pGP approach yields \textit{the best performance in all four performance metrics}. In particular,

$\bullet$ In terms of $e_F$ and $e_R$, the proposed pGP outperforms the interpolation approach and the global POD approach respectively in 365 and 395 out of the 405 testing cases. 

$\bullet$ In terms of $e_A$, the proposed pGP outperforms the interpolation approach and the global POD approach in 351 and 385 out of the 405 testing cases, respectively. 

$\bullet$ In terms of $e_I$, the proposed pGP outperforms the interpolation approach and the global POD approach in 372 and 349 out of the 405 testing cases, respectively. 

\vspace{-10pt}

\begin{figure}[h!] 
\includegraphics[width=0.95\textwidth]{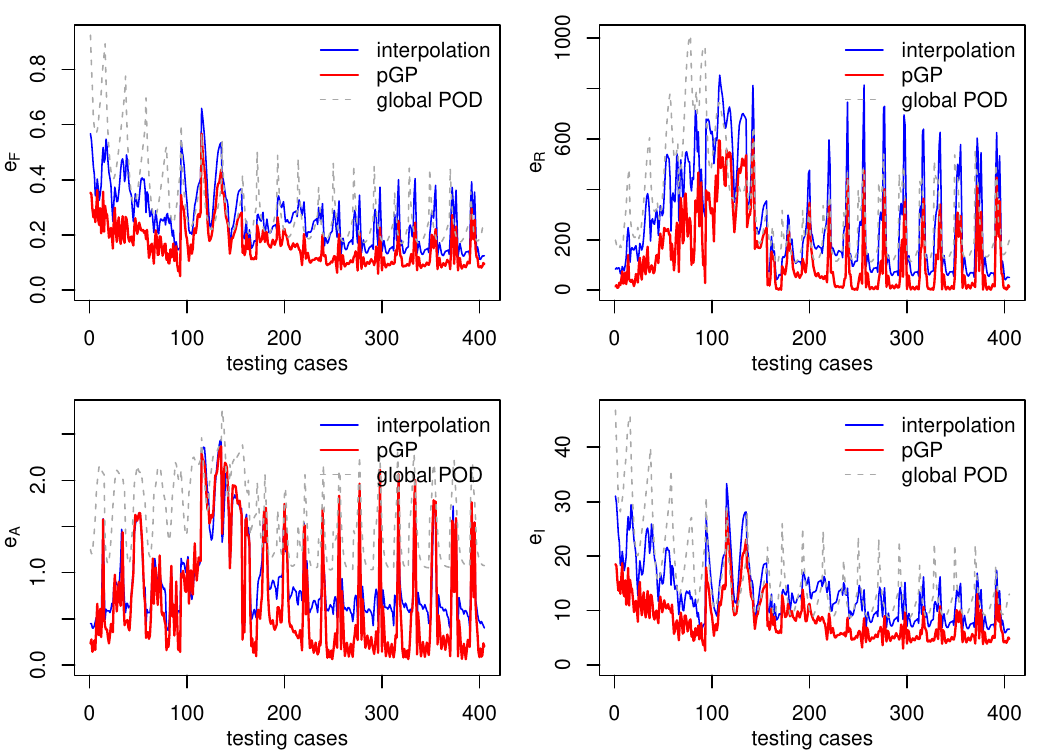}
\vspace{-8pt}
\caption{Testing errors, $e_F$, $e_R$, $e_A$ and $e_I$, for all 405 testing cases ($r=10$)}
\label{fig:error_K10}
\end{figure}

Next, we increase the reduced dimension to $r=10$, and the updated comparison results are shown in Figure \ref{fig:error_K10}. The proposed pGP still yields the best overall performance. We also note that, when the dimension of the reduced-order basis increases, all three approaches yield smaller errors in terms of $e_F$, $e_A$ and $e_I$. However, in terms of the relative error $e_R$, the errors seem to become larger. This is not surprising as further investigation shows that the first 10 optimal POD modes (obtained from performing SVD directly on the testing snapshot data) can already capture about 97\% of the total energy, making the denominator $e^*$ in (\ref{eq:e_R}) very small.


\begin{figure}[h!] 
\center
\includegraphics[width=0.6\textwidth]{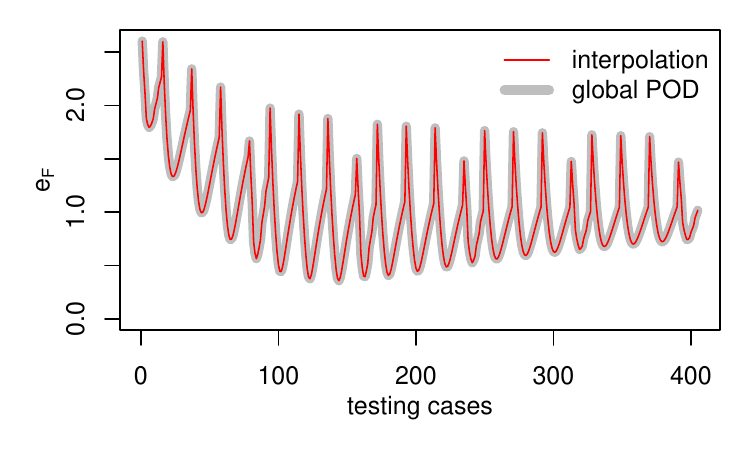}
\vspace{-12pt}
\caption{The proposed pGP includes the global POD approach as its special case by yielding almost identical $e_F$ errors for all 405 testing cases when the dimension of the reduced-order basis is increased to $r=20$.}
\label{fig:global_converge}
\end{figure}
\textbf{Relationship between the proposed and global POD approaches.}
Following the discussions above, one may wonder what if the dimension of the reduced-order basis is further increased. Note that, an important property of the proposed pGP is that it includes the use of global POD basis as its special case. As already discussed in Sec.\ref{sec:discussions}, consider the squared Exponential kernel, $\omega(\bm{\theta}_i,\bm{\theta}_j) =\xi_1^2 \delta_{\{\bm{\theta}_i=\bm{\theta}_j\}}+  \xi_2^2 \exp \{-(\bm{\theta}_i-\bm{\theta}_j)^T \bm{K}_{\omega} (\bm{\theta}_i-\bm{\theta}_j)/2\}$, where $\bm{K}_{\omega} = \text{diag}\{\xi_3^{-2},\xi_4^{-2}\}$. If the estimated $\xi_1$, $\xi_3$ and $\xi_4$ are all very small, then, $\omega(\bm{\theta}_i,\bm{\theta}_j)$ is close to zero and it is implied by (\ref{eq:pred_mean})  that $\bm{u}_*=\bm{0}$ and $p_*|\bm{p}^{\text{train}} \sim \mathcal{P}_{m_*}(\mathcal{MN}(\bm{0},\tilde{\bm{K}}_{*}))$. Because the global POD basis is chosen as the reference point, any subspace (corresponding to different $\bm{\theta} \in \Theta^{\text{test}}$) on the Grassmann manifold can be essentially seen as the \textit{i.i.d.} samples from a pGD, $\mathcal{P}_{m_*}(\mathcal{MN}(\bm{0},\tilde{\bm{K}}))$. Hence, the point prediction, $\hat{p}_*$, is close to the subspace spanned by the global POD basis. This theoretical result is illustrated in Figure \ref{fig:global_converge}, which shows that the proposed pGP yields almost identical $e_F$ errors as the global POD approach does for all 405 testing cases when the dimension of the reduced-order basis is increased to $r=20$.

In the Appendix, we provide one more example that shows how the proposed pGP includes the use of global POD as its special case. 

\vspace{-48pt}
\begin{figure}[h!] 
\center
\includegraphics[width=0.7\textwidth]{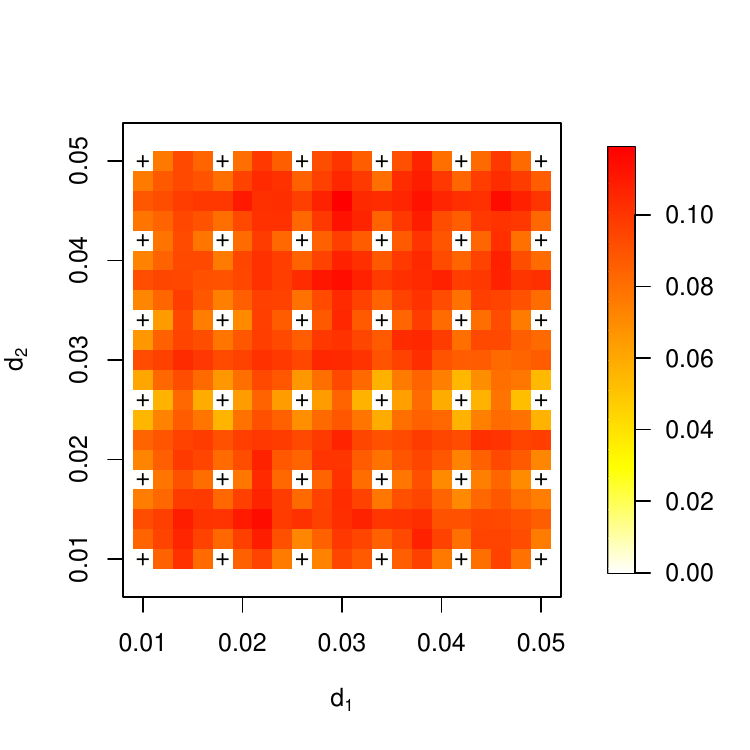}
\vspace{-18pt}
\caption{Bootstrapping standard deviation of the predicted subspace over the parameter space: the uncertainty becomes larger for parameter settings away from training points indicated by ``$+$''.}
\label{fig:UQ}
\end{figure}
\textbf{Statistical uncertainty quantification.} As a statistical learning approach, the proposed pGP is capable of quantifying the statistical uncertainty of the predicted POD basis using a simple simulation-based procedure: \textit{i}) sample a large number of vectors $\bm{y}_*^{(i)}$, $i=1,2,\cdots,\tilde{n}$, from the multivariate Gaussian distribution $\mathcal{MN}(\bm{u}_*,\tilde{\bm{K}}_{*})$ where $\bm{u}_*$ and $\tilde{\bm{K}}_{*}$ are respectively defined in (\ref{eq:pred_mean_u}) and (\ref{eq:pred_mean_K}); \textit{ii}) for each generated $\bm{y}_*^{(i)}$, compute the corresponding POD basis $\bm{\Phi}_*^{(i)}$ which spans the subspace $p_*^{(i)}$ (in other worlds, the first two steps perform the random subspace generation from the distribution $\mathcal{P}_{m_*}(\mathcal{MN}(\bm{u}_*,\tilde{\bm{K}}_{*}))$ obtained from (\ref{eq:pred_mean})); \textit{iii}) for each subspace $p_*^{(i)}$, compute the principal angle (i.e., deviation) between the sampled $p_*^{(i)}$ and the predicted subspace $p_*$, and we denote such a distance by $\zeta^{(i)}$; and \textit{iv}) compute the bootstrapping standard deviation of $\{\zeta^{(i)}\}_{i=1}^{\tilde{n}}$, which is used to measure the uncertainty associated with the predicted POD basis at a given parameter settings. 

To illustrate the output generated by the procedure above, Figure \ref{fig:UQ} shows the bootstrapping standard deviation of the predicted subspace at each testing parameter setting over the parameter space (here, $\tilde{n}=1000$). In this figure, the small cross ``$+$'' indicates the training points where there is no uncertainty. A clear pattern is observed: the uncertainty is the lower for parameter settings closer to the training points, while the uncertainty becomes larger for parameter settings away from training points. In general, the uncertainty is the largest halfway between existing training points. Such a pattern is consistent with our intuition.

\textbf{Parameter Selection Capabilities.} As discussed in Sec.\ref{sec:discussions}, it is often the case that not all parameters are equally important in terms of their predictive power for POD basis modes. In other words, the POD modes can be sensitive to some parameters while independent of the others. Note that, this idea may seem to share some similarities with the concept of active subspace \citep{Constantine2014}. The difference between the concept of active subspace and parameter selection discussed here is that, the former identifies a low-dimensional space of model input which contains the detected directions of the strongest variability in model output, while the latter refers to the process of identifying the parameters which have the strongest predictive power of the POD modes in a statistical model (for a specific problem and given a specific training dataset). 

To demonstrate the parameter selection capabilities, we expand the parameter space and consider the change of both the velocity $\bm{v}=(v_1,v_2)$ and diffusivity $\bm{D}=\text{diag}\{d_1,d_2\}$. In particular, the training data are simulated for $(v_1,v_2)$ values from a 2D mesh grid with uniformly spaced x- and y-coordinates in the interval [-0.1, 0.1] with a step size of 0.05, and for $(d_1,d_2)$ values  from a 2D mesh grid with uniformly spaced x- and y-coordinates in the interval [0.01, 0.05] with a step size of 0.05. 
Before the pGP model is trained, the values of each of the four parameters are standardized between 0 an 1, and the standardized parameter space is $[0,1]^4$. 

In \cite{Rasmussen2005}, the squared Exponential kernel is considered when introducing the parameter selection problem for GP. In our experiment, we adopt the same type of kernel as follows with a minor modification:
\begin{equation}
\omega(\bm{\theta}_i,\bm{\theta}_j) =\xi_1^2 \delta_{\{\bm{\theta}_i=\bm{\theta}_j\}}+  \xi_2^2 \exp \left\{-\frac{(\bm{\theta}_i-\bm{\theta}_j)^T \bm{K}_{\omega} (\bm{\theta}_i-\bm{\theta}_j)}{2} \right\}. 
\end{equation}
where $\bm{\theta}=(v_1,v_2, d_1,d_2)$,  $\bm{K}_{\omega} = \text{diag}\{\xi_3^{-2},\xi_3^{-2}, \xi_4^{-2},\xi_4^{-2}\}$, and $(\xi_1,\xi_2, \xi_3, \xi_4)$ are the parameters in the kernel function to be estimated from data. 
Note that, $\xi_3$ and $\xi_4$ are known as the \textit{characteristic length-scales} for velocity and diffusivity parameters. Following the idea of automatic relevance determination, a smaller length-scale indicates a stronger effect of a parameter on the prediction \citep{neal2012bayesian}. One may also note that  we let $v_1$ and $v_2$ share the same length-scale $\xi_3$, and let $d_1$ and $d_2$ share the same length-scale $\xi_4$. This set up is appropriate considering the parameter mesh grids used to generate the training data, and there is no strong reason to justify why $v_1$ and $v_2$ (or, $d_1$ and $d_2$) should have very different length-scales.

\vspace{-38pt}
\begin{figure}[h!] 
\center
\includegraphics[width=0.75\textwidth]{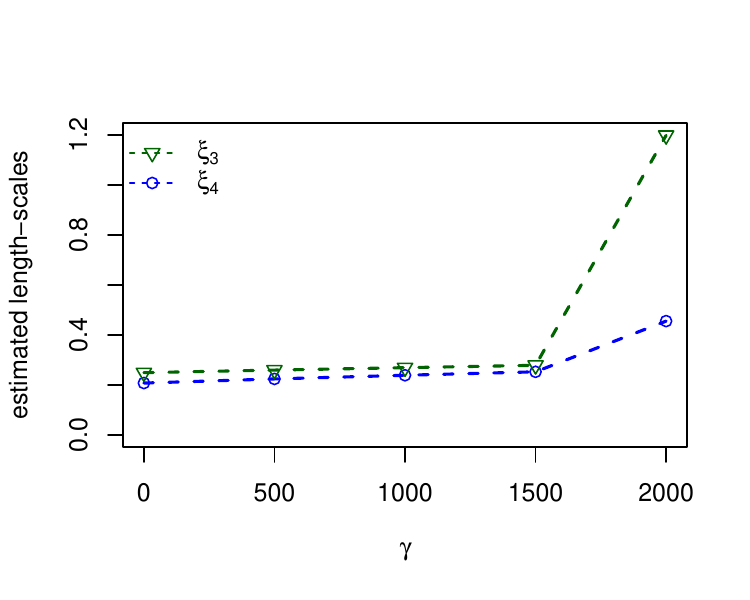}
\vspace{-18pt}
\caption{Growth of the estimated  length-scales $\hat{\xi}_3$ and $\hat{\xi}_4$ as the tuning parameter $\gamma$ increases.}
\label{fig:parameter_selection}
\end{figure}

The parameters $(\xi_1,\xi_2, \xi_3, \xi_4)$ are estimated by the maximum likelihood method \citep{Rasmussen2005}. In particular, to demonstrate the parameter selection capabilities, an $\ell_2$ regularization is imposed on $\xi_3$ and $\xi_4$, i.e., $-\gamma(\xi_3^{-2} + \xi_4^{-2})$ for some tuning parameter $\gamma \geq 0$. Note that, (i) the regularization is only imposed on the length-scales $\xi_3$ and $\xi_4$, and (ii)
as $\gamma$ increases, the regularization tends to make both $\xi_3$ and $\xi_4$ larger, i.e., to make the parameters (features) less influential in the regression model. Hence, the length-scale associated with the parameter that has less predictive power is expected to grow faster, helping us identify the important parameters. 

Figure \ref{fig:parameter_selection} shows the estimated  length-scales $\hat{\xi}_3$ and $\hat{\xi}_4$ for five different values of $\gamma$, i.e., 0, 500, 1000, 1500 and 2000. It is seen that, (i) when $\gamma=0$ (i.e., no regularization is added), the estimated values of $\xi_3$ and $\xi_4$ are respectively $\hat{\xi}_3=0.249$ (length-scale associated with the velocity parameter) and $\hat{\xi}_4=0.208$ (length-scale associated with the diffusion parameter). Because $\hat{\xi}_3$ is slightly larger than $\hat{\xi}_4$, indicating that the diffusion parameter may have a slightly higher predictive power than the velocity parameter; (ii) when $\gamma$ increases, both  $\hat{\xi}_3$ and $\hat{\xi}_4$ become larger as expected; (iii) when $\gamma=2000$, the estimated length scale $\hat{\xi}_3=1.199$ is about 2.6 times higher than $\hat{\xi}_4=0.456$, clearly indicating that the diffusion parameter has a higher predictive power than the velocity parameter over the parameter space considered in this experiment. The observations above demonstrate the parameter selection capabilities of the proposed pGP approach when constructing a predictive model for the POD modes.


\vspace{6pt}
\section{Conclusions}

This work successfully demonstrated the great potential of introducing machine learning to the field of robust reduced-order modeling. A supervised learning problem, based on the proposed pGP regression, has been developed to predict the optimal POD basis. The proposed approach leverages an injective mapping from the parameter space to the Grassmann manifold that contains the optimal vector subspaces. The condition for such a mapping to be injective has been obtained, and sufficient technical details were presented for predicting the POD basis using the proposed pGP regression. Numerical experiments showed that the proposed pGP approach not only improves the accuracy of ROM against parameter change, but also enables uncertainty quantification and parameter selection, potentially reducing the dimension of the parametric space when adapting the POD basis. We would also like to point out that there exists a critical link between the proposed pGP and statistical experimental design/reinforcement learning that enable sequential exploration of a system's behavior in a reduced parameter space. This link is expected to create important future research trajectories, potentially attracting more researchers into this field and making tangible contributions to Scientific Machine Learning research and its applications. Code and data are available on GitHub: 
\url{https://github.com/dnncode/pGP}. 

\bibliographystyle{siamplain}
\bibliography{references}

\vspace{20pt}
\appendix 

\section{Heat process in powder bed fusion} 

In the Appendix, we provide one more example that shows how the proposed pGP includes the use of global POD basis as its special case, and yields more accurate predictions of the POD basis than interpolation. 

\begin{figure}[h!]
    \centering
    \includegraphics[width=0.85\linewidth]{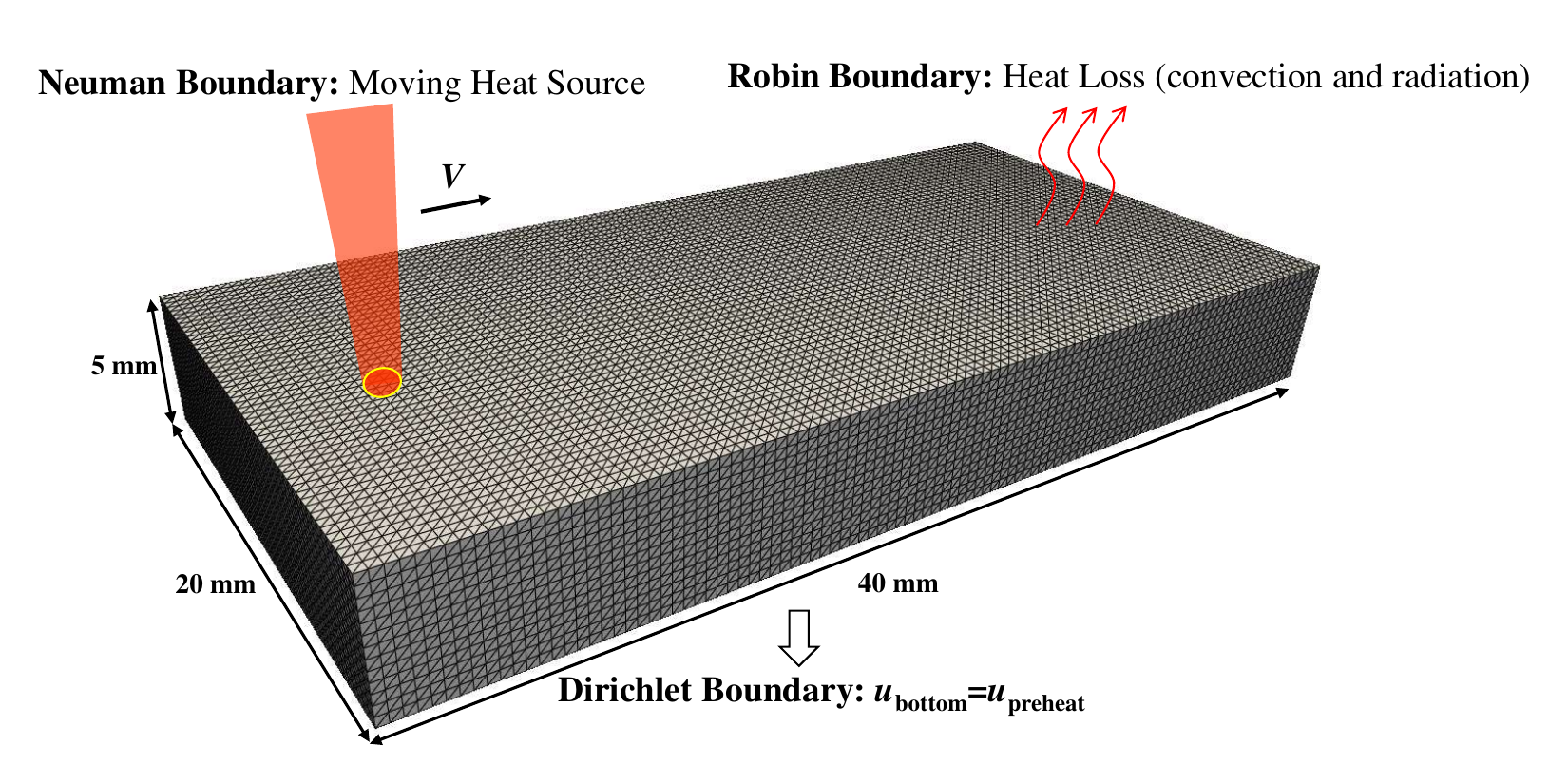}
    \vspace{-8pt}
    \caption{Illustration of problem setup, spatial domain and boundary conditions.}
    \label{fig:problem}
\end{figure}

\textbf{Experiment setup}. 
We focus on the transient heat transfer in the single-track L-PBF (laser Powder Bed Fusion) process as shown in Figure \ref{fig:problem}. PBF with metals or polymers is one of the most common 3D printing technologies in industrial additive manufacturing (AM). PBF works by jointing powdered material point by point using energy sources such as lasers or electron beams. Thermal analysis is essential for product quality control which is key to spreading laser-based AM techniques  \citep{yan2018review} and manufacturing process design  \citep{liao2022hybrid}. 

The vacuum chamber is preheated to a specific temperature and the spatial domain is a combination of the solid substrate and the powder layer. 
The transient heat transfer process is governed by a PDE: 
\begin{equation}
    \rho c_p \frac{\partial u(x,t)}{\partial t} - \nabla \cdot k\nabla u(x,t) = 0, \quad (x,t)\in \Omega\times[0,t_f],
\label{e:PDE heat transfer}
\end{equation}
where $\rho$ denotes the density, $c_p$ is the specific heat capacity, $k$ is the thermal conductivity of the material, $u(x,t)$ represents the desired temperature over a space-time domain $\Omega\times[0,t_f]$. 
The heat flux boundary condition in PBF manufacturing processes is described as
\begin{equation}
    (-k\nabla u(x,t))\cdot \mathbf{n}=q_h + q_c + q_r,\quad (x,t)\in\partial\Omega\times[0,t_f],
\label{e:boundary}
\end{equation}
where $\mathbf{n}$ is the (outward) unit normal vector to the surface, $q_h$, $q_c$, and $q_r$ are the heat flux due to the heat source, convection and radiation, respectively. 

\vspace{-10pt}

A moving Gaussian heat source considered as follows:
\begin{equation}
    q_h = -\frac{2\alpha \tau}{\pi \gamma^2}\text{exp}(-\frac{2\|x-x^c(t)\|^2}{\gamma^2}), \quad x \in \partial\Omega_t,
\end{equation}
where $\partial\Omega_t$ denotes the top surface of domain, $\alpha$ is the heat absorptivity, $\tau$ is the laser power, $\gamma$ is the effective laser (or electron) beam radius, $x^c(t)$ is the center location of the energy beam at time $t$. 

The convection and radiation, for $x \in \partial\Omega_t\cup \partial\Omega_s$, are modeled as
\begin{equation}
q_c(x,t)=h(u(x,t)-u_a), \quad q_r(x,t)=\sigma_s\epsilon(u(x,t)^4-u_a^4), 
\end{equation}
where $\partial\Omega_s$ denotes the side surfaces of domain, $h$ is the heat convection coefficient, $u_a$ is the ambient temperature, $\sigma_s$ is the Stefan-Boltzmann constant, $\epsilon$ is the emissivity of the material. The bottom surface of domain is defined with a Dirichlet boundary, $u(x,\cdot)=u_b$,  
where $x\in\partial\Omega_b$, $\partial\Omega_b$ denotes the bottom surface of domain and $u_b$ is the fixed temperature. 
Finally, the initial condition is defined as $u(\cdot,0)=u_0$
where $x\in\Omega$ and $U_0$ is the initial temperature. Note that, $\partial\Omega = \partial\Omega_t \cup \partial\Omega_s \cup \partial\Omega_b$.

The finite element method is used to solve the governing equations. 
Multiplying both sides of (\ref{e:PDE heat transfer}) by a test function $\phi$ and integrating over $\Omega$ gives
\begin{equation}
    \int_{\Omega}\rho c_p \frac{\partial u(x,t)}{\partial t}\phi dV - \int_{\Omega}\nabla \cdot k\nabla u(x,t)\phi dV = 0, \quad (x,t)\in \Omega\times[0,t_f],
\label{e:weak formulation}
\end{equation}
The solution $u(x,t)$ and the test function are assumed to belong to Hilbert spaces. Unlike the original formulation that (\ref{e:PDE heat transfer}) holds for all points in $\Omega$, the weak formulation requires (\ref{e:weak formulation}) to hold for all test functions in test function space. The Galerkin method assumes that the solution $u(x,t)$ belongs to the same Hilbert space as the test functions. According to Green's first identity, (\ref{e:weak formulation}) gives
\begin{equation}
    \int_{\Omega}\rho c_p \frac{\partial u(x,t)}{\partial t}\phi dV + \int_{\Omega} k\nabla u(x,t)\cdot\nabla\phi dV + \int_{\partial\Omega}(-k\nabla u(x,t))\cdot\mathbf{n}\phi dS = 0,
\label{e:weak formulation green}
\end{equation}
Then, the discretization by Galerkin method is applied to the weak formulation and the approximated solution $u_d(x,t)$ is given by $u_d(x,t)=\sum_{i=1}^{N}u_i(t)\psi_i(x)$, where $N$ is the number of nodes. For every test function $\psi_j(x)$, we have
\begin{equation}
    \rho c_p \sum_{i=1}^{N}\frac{\partial u_i(t)}{\partial t}\int_{\Omega}\psi_i\psi_jdV + k\sum_{i=1}^{N}u_i(t)\int_{\Omega}\nabla\psi_i\nabla\psi_jdV + \sum_{i=1}^{N}\int_{\partial\Omega}(-k\nabla u_i(t)\psi_i)\cdot\mathbf{n}\psi_j dS=0
\label{e:discretization}
\end{equation}




\vspace{-12pt}
\begin{figure}[h!]
    \centering
\includegraphics[width=0.9\linewidth]{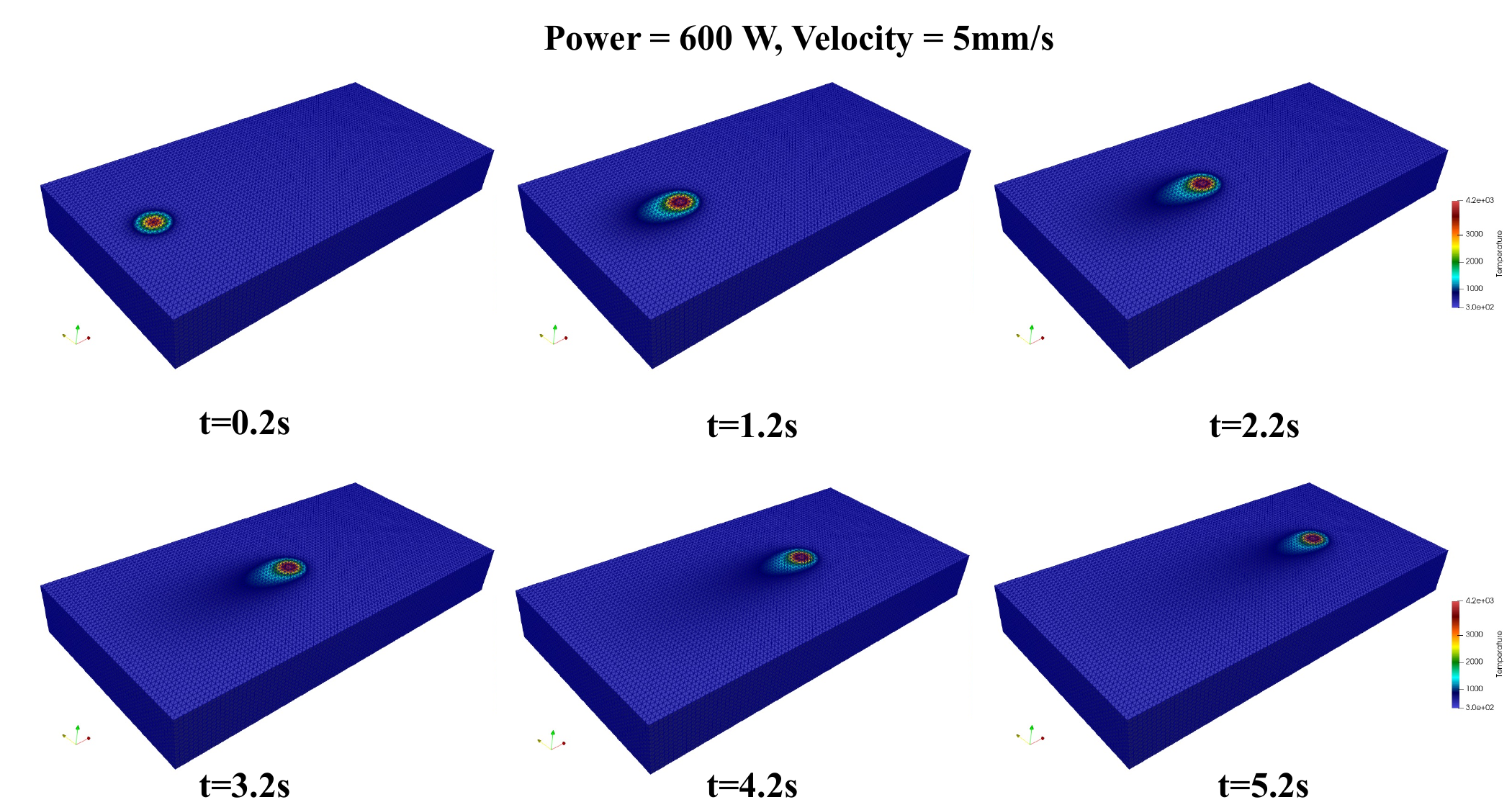}
    \vspace{-8pt}
    \caption{Temperature field at 600W laser power with the heat source moving at a speed of 5mm/s.}
    \label{fig:temperature}
\end{figure}

We generate data for 7 laser power levels (in Watt): $\{500,550,600,650,700,750,800\}$, using the finite element method. As an illustration, Figure \ref{fig:temperature} shows the temperature field at selected times. Since seven laser power levels are used to generate the data, the Leave-One-Out Cross-Validation (LOOCV) is performed to compare the model performance. We let $r=10$ and $m(\tau)$ be a constant basepoint which is the subspace spanned by the global POD basis. The kernel $\omega$ is chosen as $\omega(\tau_i,\tau_j)=\xi_1\exp(-\xi_2^{-1}|\tau_i-\tau_j|)$, and $\bm{K}=\sigma_{\bm{K}}\bm{I}$.

\begin{figure}[h!] 
\centering
\includegraphics[width=0.8\textwidth]{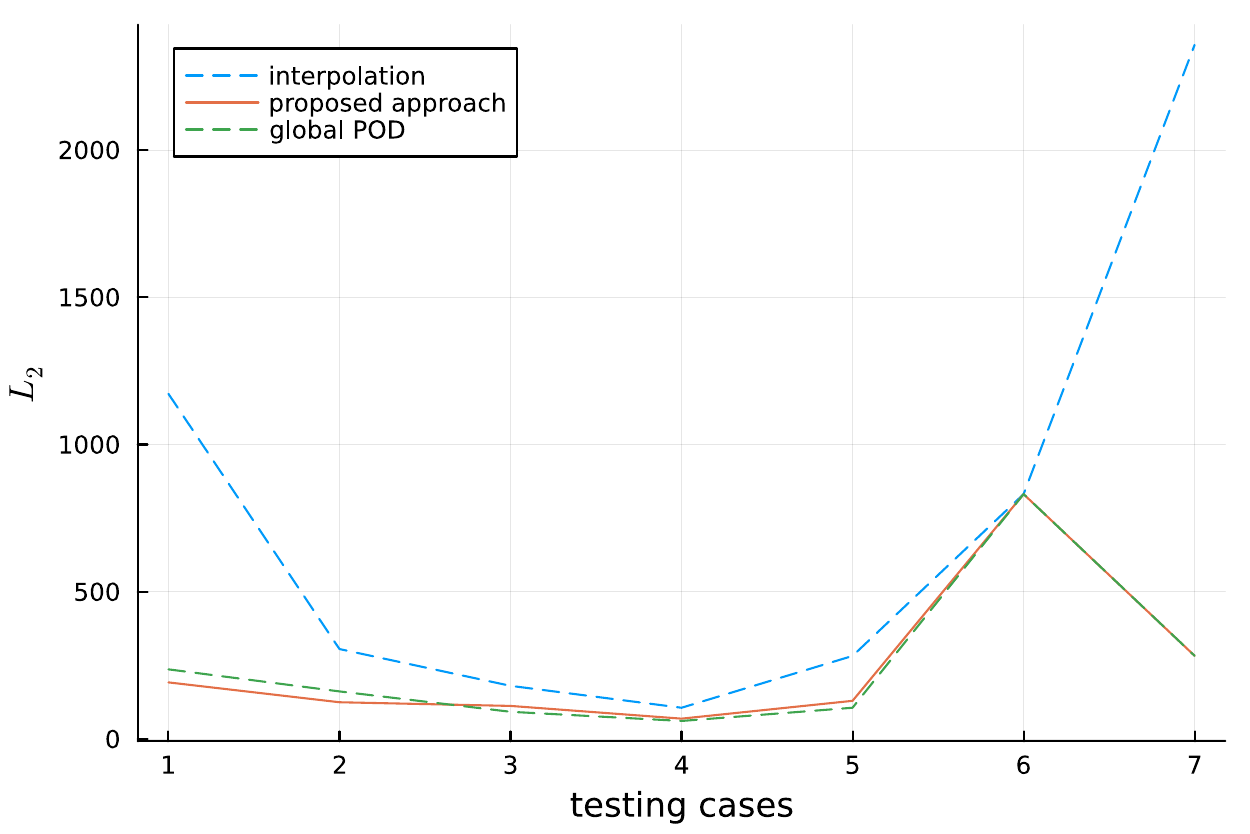}
\caption{Leave-One-Out Cross-Validation (LOOCV) errors.}
\label{fig:am_error}
\end{figure}

\textbf{Experiment Results.} Figure \ref{fig:am_error} shows LOOCV testing $\ell_2$ errors for the proposed approach, the existing POD interpolation method, and the use of global POD bases. It is important note that, 

$\bullet$ The use of global POD basis and the proposed pGP approach yield better and very similar results. 

$\bullet$ More importantly, the proposed pGP is able to provide an explanation to such an observation. In short, this is because the proposed pGP (\textit{i}) involves the use of global POD basis as its special case (due to its parameter selection capabilities), and (\textit{ii}) successfully identifies that the POD basis, in this example, is not sensitive to the change of parameters, and the use of the global POD basis becomes appropriate. 

\vspace{-12pt}
To elaborate, the kernel $\omega$ is chosen as $\omega(\tau_i,\tau_j)=\xi_1\exp(-\xi_2^{-1}|\tau_i-\tau_j|)$ in this example. When the LOOCV is performed, we note that the estimated values for $\xi_2$ ranges between 40 to 80, effectively making $\omega(\bm{\theta}_i,\bm{\theta}_j)$ close to zero. In this case, (\ref{eq:pred_mean}) immediately implies that $\bm{u}_*=\bm{0}$ and $p_*|\bm{p}^{\text{train}} \sim \mathcal{P}_{m_*}(\mathcal{MN}(\bm{0},\tilde{\bm{K}}_{*}))$. Because the global POD basis is chosen as the reference point, then, any subspace (corresponding different power levels $\tau$) on the Grassmann manifold can be seen as \textit{i.i.d.} samples from a pGD, $\mathcal{P}_{m_*}(\mathcal{MN}(\bm{0},\tilde{\bm{K}}))$. Hence, the point prediction, $\hat{p}_*$, is close to the subspace spanned by the global POD basis, justifying the observation in Figure \ref{fig:am_error}.



\end{document}